\documentclass{article}
\usepackage[submission]{colm2026_conference}

\usepackage{hyperref}
\usepackage{url}
\usepackage{booktabs}
\usepackage{lineno}
\usepackage{color}
\usepackage{tcolorbox}
\usepackage{caption}
\usepackage{subcaption}
\usepackage{amssymb}
\usepackage[T1]{fontenc}
\tcbuselibrary{skins}

\usepackage[table]{xcolor}
\usepackage{graphicx}
\usepackage{tabularx}
\usepackage{amsmath}
\usepackage[capitalize]{cleveref}
\usepackage{arydshln}
\usepackage{mathtools}
\usepackage{multirow}
\usepackage{lmodern}
\usepackage{setspace}
\hypersetup{colorlinks,linkcolor={blue},citecolor={blue},urlcolor={blue}}

\newcommand{\myalg}{\textsc{VAEX-Bench}}

\definecolor{absgray}{RGB}{242,243,245}
\definecolor{metablue}{RGB}{0,102,204}

\newcommand{\customabstractpage}{
\begin{tcolorbox}[
    enhanced,
    colback=absgray,
    colframe=absgray,
    boxrule=0pt,
    arc=8pt,
    left=3mm,
    right=3mm,
    top=3mm,
    bottom=3mm
]

{\Large\bfseries
Reasoning over Video: Evaluating How MLLMs Extract, Integrate, and Reconstruct Spatiotemporal Evidence
\par}

\vspace{3mm}

Seunghwan Bang$^{1,*}$ and Hwanjun Song$^{2}$\par

\vspace{1mm}

$^{1}$UNIST, $^{2}$KAIST \par
* This work was done while interning at KAIST.
\vspace{4mm}

\noindent
 The growing interest in embodied agents increases the demand for spatiotemporal video understanding, yet existing benchmarks largely emphasize extractive reasoning, where answers can be explicitly presented within spatiotemporal events. It remains unclear whether multimodal large language models can instead perform abstractive spatiotemporal reasoning, which requires integrating observations over time, combining dispersed cues, and inferring implicit spatial and contextual structure. To address this gap, we formalize abstractive spatiotemporal reasoning from videos by introducing a structured evaluation taxonomy that systematically targets its core dimensions and constructs a controllable, scenario-driven synthetic egocentric video dataset tailored to evaluate abstractive spatiotemporal reasoning capabilities, spanning object-, room-, and floor-plan–level scenarios. Based on this framework, we present \myalg{}, a benchmark comprising five abstractive reasoning tasks together with their extractive counterparts. Our extensive experiments compare the performance of state-of-the-art MLLMs under extractive and abstractive settings, exposing their limitations on abstractive tasks and providing a fine-grained analysis of the underlying bottlenecks. The dataset will be released soon.

\vspace{4mm}

\noindent
\begin{minipage}[t]{0.65\textwidth}
{\small
\textbf{Date:} March 6, 2026 \par
\textbf{Correspondence:} Hwanjun Song at {\color{metablue}\href{mailto:songhwanjun@kaist.ac.kr}{songhwanjun@kaist.ac.kr}} \par
\textbf{First Author:} Seunghwan Bang at {\color{metablue}\href{mailto:hwani.choi@kaist.ac.kr}{bang1422@gmail.com}} \par
\textbf{Project Page:} {\color{metablue}\url{https://disl-lab.github.io/VAEX-Bench/}}
}
\end{minipage}
\hfill
\begin{minipage}[t]{0.27\textwidth}
\vspace*{-0.1cm}
\raggedleft
\includegraphics[width=1.0\linewidth]{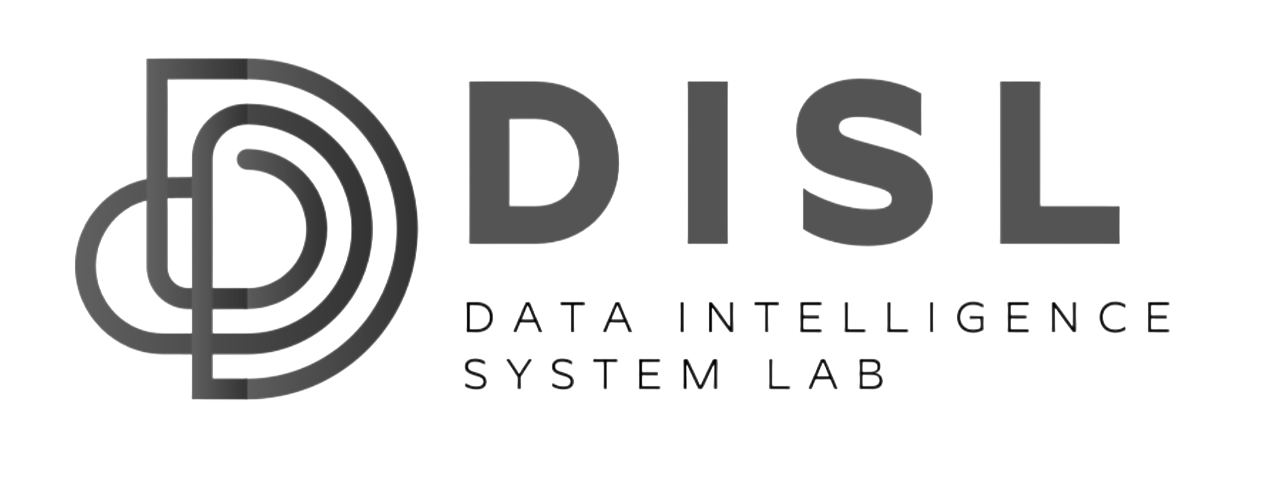}
\end{minipage}

\end{tcolorbox}
}

\begin{document}

\thispagestyle{empty}

\vspace*{-1.3cm}
\customabstractpage

\setcounter{section}{-1}
\section{Overview}
\begin{figure}[htbp]
    \centering
    \hspace*{0.7cm}
    \includegraphics[width=10cm]{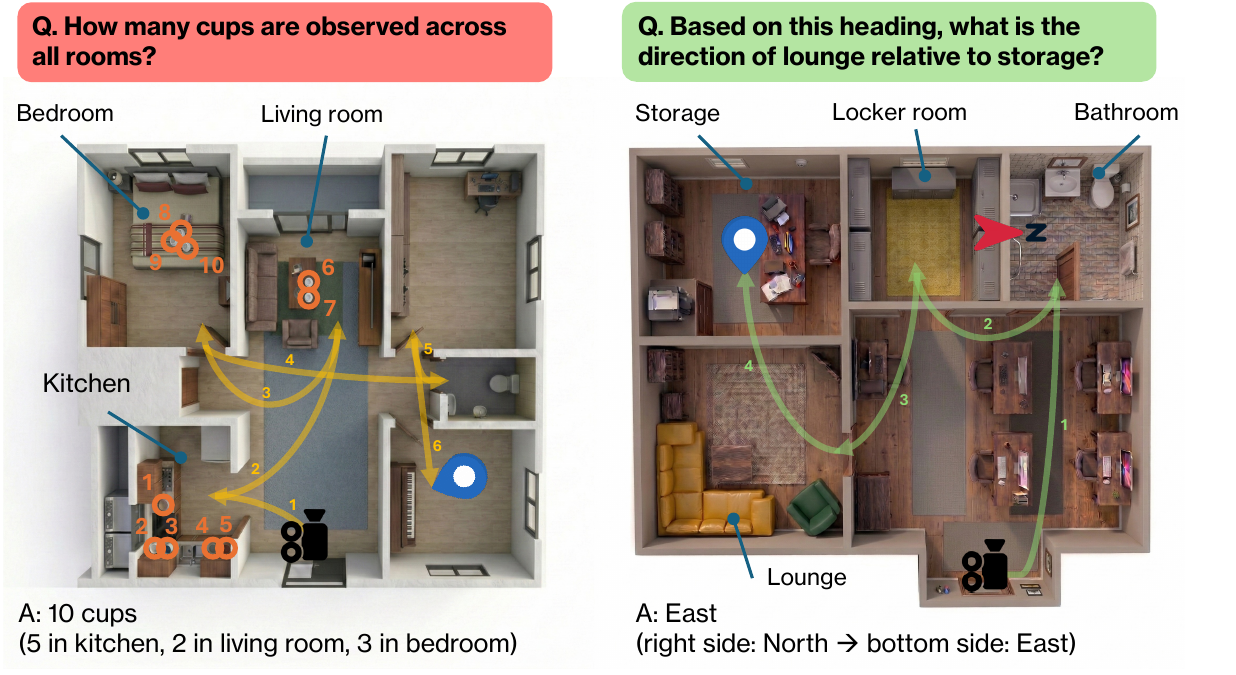}
    \captionof{figure}{Two examples of abstractive spatiotemporal queries from \myalg{}: (Left) Global Counting, requiring aggregation of objects observed across the traversal; (Right) Map Direction, requiring inference of relative spatial directions between rooms.}
    \label{fig:teaser}
\end{figure}
\section{Introduction}\label{sec:introduction}

The rapid rise of embodied agents has fundamentally reshaped the demands placed on video understanding systems \citep{castro2020lifeqa, jang2017tgif, wu2024longvideobench}. Moving beyond static image perception and short-clip recognition, modern multimodal large language models (MLLMs) are increasingly expected to reason over long-horizon egocentric video streams \citep{xu2024survey, sermanet2024robovqa, wang2025internvl3, comanici2025gemini, hong2025glm}, where observations are fragmented, viewpoints shift continuously, and critical evidence may be temporally dispersed. Thus, recent video spatiotemporal benchmarks, such as VSI-Bench \citep{yang2025thinking} and VSTI-Bench \citep{fan2025vlm}, have expanded evaluation toward \emph{spatiotemporal} reasoning by emphasizing the integration of visual cues across time and three-dimensional (3D) spatial relations. This evolution reflects a broader transition from identifying ``what is visible in a frame?'' to ``what can be inferred from an evolving first-person experience,'' a capability that lies at the core of embodied intelligence.

Despite this progress, existing video spatiotemporal benchmarks \citep{yang2025thinking, fan2025vlm} have primarily been designed to evaluate whether MLLMs can correctly identify objects, actions, or relations that are explicitly present in spatiotemporal events (\emph{e.g.}, inferring object appearance order, identifying the nearest object), a setting we refer to as \emph{extractive} spatiotemporal reasoning. Although such evaluation is essential for measuring perceptual grounding and local spatiotemporal recognition \citep{kesenvilma, ning2025video}, it does not directly assess whether MLLMs can form a coherent global representation of the environment or synthesize observations into a structured understanding of the depicted world \citep{wang2024sok, cheng2025v, zhou2025vlm4d}.

Building on this gap, we move toward \emph{abstractive} spatiotemporal reasoning, which depends on constructing and maintaining a globally integrated spatiotemporal world model to enable inference. That is, rather than reasoning about locally observable relations from a single viewpoint, MLLMs must synthesize fragmented egocentric observations into a coherent cognitive map that supports global orientation and aggregation under partial observability in Figure \ref{fig:teaser} (\emph{e.g.}, reconstructing the underlying floor plan, counting objects across rooms). Therefore, this shift calls for a principled benchmark designed to jointly evaluate both extractive and abstractive spatiotemporal reasoning.

To realize this, we depart from existing video benchmark construction \citep{cheng2025v, rodin2025easg, wang2025lvbench, yang2025thinking, fan2025vlm}, where questions are annotated on top of pre-captured real-world videos (\emph{e.g.}, ARKitScenes \citep{baruch2021arkitscenes}, ScanNet++ \citep{yeshwanth2023scannet++}). As the visual evidence is fixed by the captured scenes and trajectories, this \emph{post-hoc} process provides limited control over scene layout, object placement, or trajectory structure, making it difficult to systematically target abstractive spatiotemporal reasoning.
Instead, we systematically expand five representative extractive task types in prior works \citep{yang2025thinking, fan2025vlm} into their abstractive counterparts under a one-to-one expansion principle (see Table \ref{tab:task_expansion}). To enable the construction of such queries, we represent environments through a structured indoor scenario factorization over ``object configuration,'' ``room composition,'' and ``floor-plan topology." This factorization allows us to explicitly specify the spatial structure and object distribution of the environment before video generation, making it possible to design reasoning queries that require integrating evidence across rooms and time.

\begin{figure}[t]
    \centering
    \includegraphics[width=0.8\textwidth]{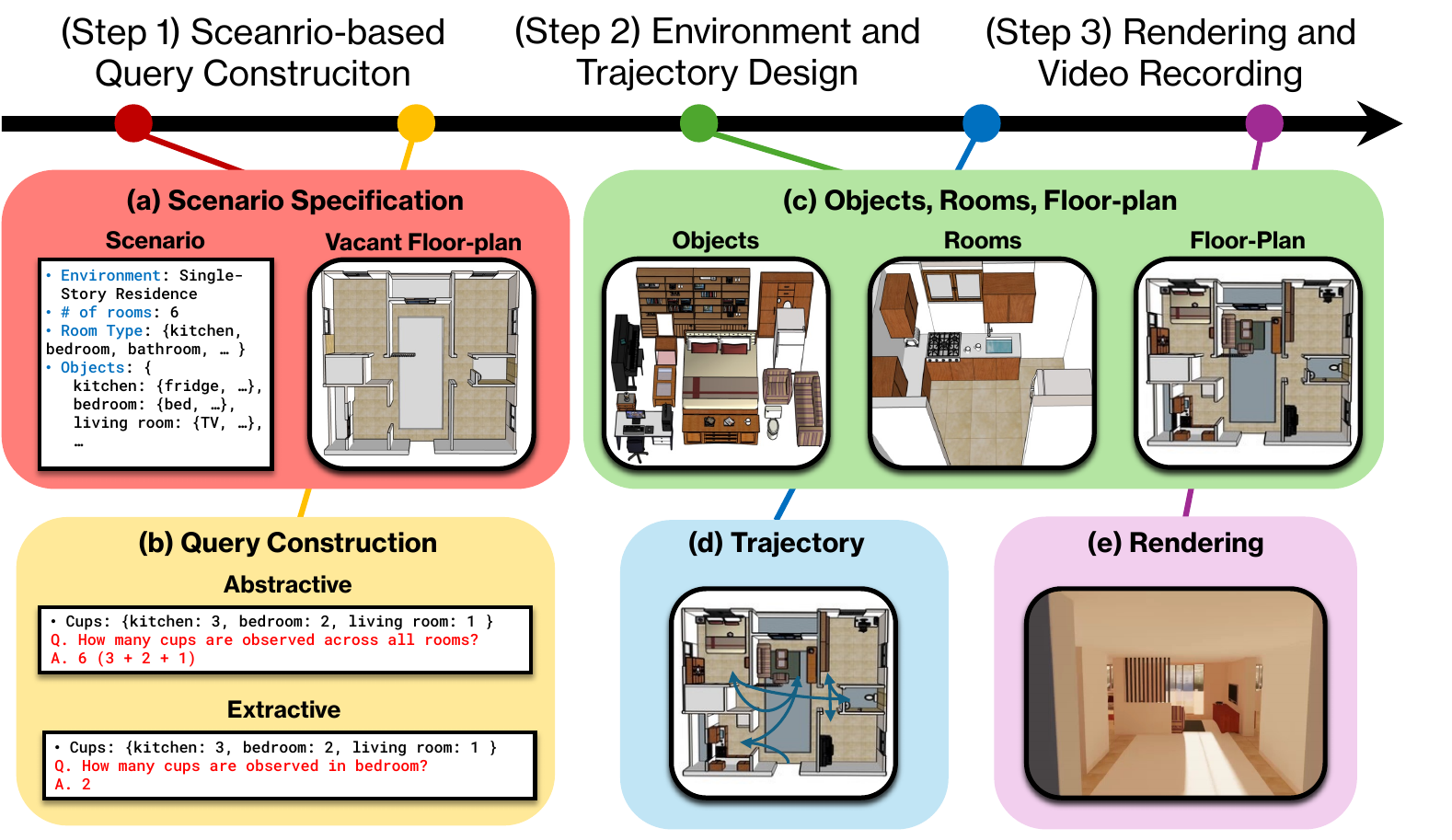}
    \caption{Query-conditioned video construction pipeline: (Step 1) Scenario-based Query Construction, followed by (Step 2) Environment and Trajectory Design and (Step 3) Rendering and Video Recording.}
    \label{fig:pipeline}
\end{figure}

As illustrated in Figure \ref{fig:pipeline}, we implement a \emph{query-conditioned} video construction pipeline. Human annotators first design reasoning queries from the scenario specification to determine what evidence must appear in the environment. The scenes are then instantiated in \texttt{SketchUp}\footnotemark[1] with specified room layouts and object placements, rendered using \texttt{Enscape}\footnotemark[2], and recorded as egocentric videos along scripted trajectories to ensure spatial and temporal consistency. After video generation, the queries are human-verified against the videos to confirm that each answer is unambiguously grounded in the visual evidence. This pipeline provides explicit control over scene structure and ground-truth relations, enabling systematic evaluation of both extractive and abstractive reasoning.

Building on this design, we construct a novel benchmark named \myalg{} (\underline{V}ideo \underline{A}bstractive and \underline{EX}tractive \underline{Bench}mark), comprising 10 controlled egocentric indoor scenarios, each paired with five abstractive reasoning tasks along with their extractive counterparts. 
In particular, as summarized in Table \ref{tab:task_expansion}, the abstractive tasks evaluate capabilities such as long-horizon memory, global orientation, metric reasoning, structural reconstruction, and trajectory-level aggregation, requiring models to integrate dispersed observations across the video. By contrast, the extractive tasks evaluate recognition of objects, actions, and relations that are explicitly observable in the video. 

Our benchmarking over 14 SOTA proprietary and open-source MLLMs reveals previously underexplored findings:
\noindent(1) Performance drops substantially when moving from extractive to abstractive reasoning across both open-source and proprietary models; \noindent(2) Accuracy is higher on multiple-choice questions than on free-form generation, indicating reliance on option cues; (3) Models exhibit unstable object perception and counting, often missing or miscounting entities across the video; (4) Models struggle to maintain long-horizon temporal memory when evidence is dispersed across the video; and (5) Models frequently fail to reconstruct globally consistent spatial layouts from egocentric observations.

\section{Related Works}\label{sec:related_works}

\noindent\textbf{Video Reasoning Benchmark.} Video reasoning benchmarks has progressed from clip comprehension to broader spatiotemporal evaluation \citep{yang2025thinking, fan2025vlm, castro2020lifeqa, tapaswi2016movieqa, jang2017tgif, fu2025video, wu2024longvideobench}. Early video benchmarks, such as MovieQA~\citep{tapaswi2016movieqa} and LifeQA~\citep{castro2020lifeqa}, primarily assessed models' ability to follow narrative structure in video content, ranging from story comprehension in curated media (\emph{e.g.}, movies,  dramas) to understanding event sequences in more natural, potentially noisy everyday footage. Subsequent works like TGIF-QA~\citep{jang2017tgif} advanced the field by shifting the focus toward more fine-grained understanding, introducing tasks that explicitly require spatiotemporal reasoning~\citep{sugandhika2025know, cheng2025v, fan2025tool, li2023discovering} to analyze repeating actions and state transitions in shorter video clips. 

\indent Most recently, VSI-Bench~\citep{yang2025thinking} extends from 2D spatiotemporal cues to 3D visual-spatial reasoning by constructing a video reasoning benchmark on indoor egocentric monocular videos captured with 3D ground-truth metadata~\citep{dai2017scannet, yeshwanth2023scannet++, baruch2021arkitscenes}, enabling evaluation of MLLMs on object-level geometric relations. Despite this progress, most existing video benchmarks remain fundamentally extractive, leaving largely untested whether models can integrate temporally dispersed visual observations into a coherent representation of the environment. To address this, we construct controllable egocentric 3D environments with scenario-driven layouts that require reconstructing the global structure of the scene from fragmented observations, enabling evaluation of long-horizon memory integration, spatial consistency, and cross-room reasoning.

\noindent\textbf{Abstractive Reasoning for LLMs.}
Recent works on LLM reasoning increasingly frames the challenge as moving beyond evidence present cue finding toward abstractive inference that integrates dispersed observations and compresses them into latent state estimates \citep{gao2025abstral, jiangrethinking, chen2025exploring, hao2024training}. Accordingly, text-based benchmarks have evaluated reasoning reliability and robustness through multi-hop question answering \citep{yang2018hotpotqa, talmor2018web, chang2022webqa}, long context integration \citep{ma2024mmlongbench, modarressi2025nolima, bai2024longbench}, causal and counterfactual inference \citep{wu2024cofca, yu2023ifqa, zhou2024causalbench}, and planning or agentic tool use \citep{liu2023agentbench, padmakumar2022teach, valmeekam2023planbench}. However, these settings largely emphasize the composition of linguistic evidence and therefore only partially capture the demands of real-world reasoning, where an agent must maintain a state under partial observability and update an internally consistent world model over space and time. Despite this, abstractive reasoning in video, the ability to reconstruct latent states and global structure beyond directly visible evidence, remains relatively understudied. Therefore, recent work broadens this notion to include long-horizon state tracking, consistency maintenance, and integration under uncertainty, motivating evaluation settings that can explicitly diagnose these capabilities in video.
\section{Taxonomy of Video Spatiotemporal Reasoning} \label{sec:taxonomy}

We distinguish two regimes of spatiotemporal reasoning from video, depending on whether answers can be derived from directly observable cues or require integrating observations to infer latent spatial and contextual structure.

\noindent\textbf{Extractive Reasoning} refers to settings where answers can be derived from directly observable cues in video frames, typically by localizing the relevant moment and reading off visible evidence. This formulation has been the practical default for prior benchmarks \cite{yang2025thinking, fan2025vlm} because it enables scalable question construction and evaluation with minimal ambiguity. For example, extractive reasoning includes tasks such as identifying the order in which objects appear (Appearance Order), determining relative directions or distances from a single viewpoint (Relative Direction or Distance), planning short navigation steps (Route Plan), or counting objects within a single room (Object Counting). However, such an evaluation can under-test the capabilities required for embodied intelligence.

\noindent\textbf{Abstractive Reasoning} instead refers to settings where answers cannot be obtained from a single observation but require integrating evidence across time to infer latent spatial and contextual structure.  To clarify how this setting differs from extractive reasoning, we define a set of additional capabilities required for such inference and organize them into a taxonomy of spatiotemporal reasoning tasks. In particular, these capabilities include retrieving evidence over long horizons (temporal memory), constructing a global spatial representation from egocentric observations (allocentric direction and distance reasoning), simulating navigation over the inferred layout, and aggregating information across multiple locations. This formulation motivates a principled expansion of extractive queries into abstractive reasoning tasks and ensures that failures can be attributed to specific reasoning bottlenecks.

\subsection{Expansion from Extractive to Abstractive Tasks}

Guided by the capability taxonomy above, our benchmark operationalizes the shift from extractive to abstractive reasoning through a one-to-one task expansion principle, as summarized in Table \ref{tab:task_expansion}. For each common extractive task (in the 2nd column), we preserve its semantic intent while removing the assumption that the decisive cue is observable at a single moment (in the 4th column). As a result, we expand five representative extractive tasks into their abstractive counterparts: Appearance Order into long-horizon memory retrieval; Relative Direction (or Distance) into allocentric spatial reasoning (or global metric estimation), Route Planning into navigation simulation, and Object Counting into global aggregation across multiple locations.

\begin{table*}[t]
    \centering
    \scriptsize
    \renewcommand{\tabularxcolumn}[1]{m{#1}}
    \begin{tabularx}{\textwidth}{@{} l @{\hspace{0.7cm}} >{\raggedright\arraybackslash}X >{\raggedright\arraybackslash}X >{\raggedright\arraybackslash}X @{}}
    \toprule
    \textbf{Category} & \textbf{Extractive} & \textbf{Abstractive} & \textbf{Expanded Capability} \\
    \midrule
    
    Chronology & 
    \textbf{Appearance Order:} \newline Order of first appearance: \textcolor{blue}{\{sofa, TV, table, cabinet\}} & 
    \textbf{Memory--Action:} \newline What activity is possible in the \textcolor{blue}{\{third\}} room? & Frame-local perception \newline $\rightarrow$ Memory-grounded perception\\
    \cmidrule{1-4}
    
    Direction & 
    \textbf{Relative Direction:} \newline Facing \textcolor{blue}{\{fridge\}} from \textcolor{blue}{\{stove\}}, where is \textcolor{blue}{\{stool\}}? & 
    \textbf{Map Direction:} \newline \textcolor{blue}{[room1 is east of room2]} \newline What is the direction of \textcolor{blue}{\{room3\}} from \textcolor{blue}{\{room4\}}? & 
    Frame-local direction \newline $\rightarrow$ Allocentric orientation \\
    \cmidrule{1-4}
    
    Distance & 
    \textbf{Relative Distance:} \newline Which object is the \newline closest to the \textcolor{blue}{\{door\}}? & 
    \textbf{Map Scale:} \newline \textcolor{blue}{[$D^2(room1, room2)=1$]} \newline What is the $D^2$ between \textcolor{blue}{\{room3\}} and \textcolor{blue}{\{room4\}}? & Frame-local proximity \newline $\rightarrow$ Global metric estimation \\
    \cmidrule{1-4}
    
    Planning & 
    \textbf{Route Plan:} \newline Facing \textcolor{blue}{\{TV\}} from \textcolor{blue}{\{sofa\}}, what sequence of turns navigate you to \textcolor{blue}{\{table\}}? & 
    \textbf{Simulation:} \newline Which room is located directly opposite to \textcolor{blue}{\{kitchen\}}? & 
    Local route planning \newline $\rightarrow$ Allocentric simulation\\
    \cmidrule{1-4}
    
    Counting & 
    \textbf{Object Counting:} \newline How many \textcolor{blue}{\{chairs\}} are in \newline the \textcolor{blue}{\{dining room\}}? & 
    \textbf{Global Counting:} \newline How many \textcolor{blue}{\{chairs\}} are observed across all rooms? & 
    Single-room counting \newline $\rightarrow$ Multi-room aggregation \\
    
    \bottomrule
    \end{tabularx}
    \caption{{Expansion from extractive to abstractive tasks.} Question examples illustrate the shift in cognitive requirements, where items \textcolor{blue}{blue} denote variable placeholders that can be adapted to different environments.}
    \label{tab:task_expansion}
\end{table*} 

\textbf{Memory--Action:} The Appearance Order task asks in what order objects appear as the camera moves and can often be solved through local recognition with limited temporal integration. We extend this to an abstractive task, where the model must recall objects encountered across rooms and infer possible actions in each space, requiring long-horizon memory across the trajectory. 

\textbf{Map Direction:} The Relative Direction task asks for the direction of an object from a single viewpoint within one room. We extend this to the abstractive task, where the model must reason about directions between rooms under a shared global orientation. This requires integrating egocentric observations to infer allocentric spatial relations across the environment.

\textbf{Map Scale:} The Relative Distance task compares distances between objects within a single scene using local perspective cues. We extend this to the abstractive task, where the model must estimate distances between rooms under a given metric reference. This requires constructing a global spatial representation from observations across the trajectory.

\textbf{Simulation:} The Route Planning task typically involves short-horizon navigation decisions based on local cues (\emph{e.g.}, turn right $\rightarrow$ go straight). We extend this to the abstractive task, where the model must identify a floor-plan hypothesis consistent with the entire traversal. This task therefore requires reasoning about the global layout rather than making step-wise navigation decisions.

\textbf{Global Counting:} The Object Counting task asks for the number of objects visible within a single room or frame. We extend this to the abstractive task, which requires tracking object occurrences across the full trajectory and aggregating them into a global total across rooms.

Note that our goal is not to replace extractive evaluation, but to jointly assess both regimes. We therefore include both abstractive and extractive tasks, enabling controlled comparisons and quantifying how performance shifts when moving from frame-grounded extraction to video abstractive reasoning.

\section{VAEX-Bench Dataset} \label{sec:benchmark}

Unlike existing extractive benchmarks \citep{yang2025thinking, fan2025vlm}, constructing benchmarks for abstractive reasoning directly from pre-collected videos is inherently challenging, as the decisive evidence required for such questions often spans multiple locations and time steps. When questions are generated after video collection, the available evidence is already fixed, making it difficult to systematically design queries that require long-horizon integration or global spatial reasoning.

To address this limitation, we propose the \emph{query-conditioned} video construction pipeline illustrated in Figure \ref{fig:pipeline}, where scenarios and questions are designed first and used to construct environments before rendering them into egocentric videos. Although this pipeline requires substantial manual effort, it enables controlled scenario construction necessary for reliable evaluation of abstractive reasoning. In particular, our pipeline is designed to satisfy three key properties: (1) Distributing decisive evidence across rooms and time to prevent purely extractive solutions; (2) Ensuring that the correct decision criteria are uniquely determined by the scene state; and (3) Varying layouts to avoid overfitting while systematically targeting different reasoning abilities.

\vspace*{-0.15cm}
\subsection{Data Construction Pipeline}
\vspace*{-0.15cm}
\smallskip\smallskip\noindent
\textbf{Step 1: Scenario-based Query Construction}

\smallskip
\noindent\underline{Scenario Specification:} Our benchmark begins with scenario specification prior to video synthesis. Each scenario defines the environment typology (\emph{e.g.}, single-story residence, multi-story residence, or office), the number and types of rooms, and the candidate objects associated with each room. These specifications are instantiated as both textual scenario descriptions and an empty floor plan in {\texttt SketchUp}\footnote{SketchUp is a widely used 3D modeling software for architectural and interior design.}, which together define room connectivity and spatial relations of the environment, as illustrated in Figure \ref{fig:pipeline}(a).

\smallskip
\noindent\underline{Query Construction:} Based on the scenario specification, human annotators design and create both extractive and abstractive queries aligned with our task suite in Table \ref{tab:task_expansion}. The questions are designed by first enumerating scenario-level facts implied by the scenario (\emph{i.e.}, room adjacency, object placements, and functional relations), and then selecting queries whose answers are uniquely determined by the resulting environment configuration, as illustrated in Figure \ref{fig:pipeline}(b). This process ensures that each query has an unambiguous answer grounded in the defined scene state. Furthermore, the queries are designed to determine what evidence must appear in the environment and how it should be distributed across the scenario, providing the foundation for constructing videos that systematically evaluate both extractive and abstractive reasoning.

\smallskip\smallskip\noindent
\textbf{Step 2: Environment and Trajectory Design}

\smallskip
\noindent\underline{Scene Construction:} Based on the constructed queries, we design the environment by finalizing the floor plan and placing objects within rooms so that the evidence required to answer each query is present in the scene. We then materialize the scene by manually creating and placing the corresponding objects in \texttt{SketchUp}, producing the complete floor plan with object inventories and placements, as illustrated in Figure \ref{fig:pipeline}(c). We then design camera trajectories that visit the necessary viewpoints and temporalize evidence acquisition, ensuring that relevant observations are revealed in a controlled order during the egocentric traversal, as illustrated in Figure \ref{fig:pipeline}(d).

\smallskip
\noindent\underline{Trajectory Constraints:}
For abstractive tasks, we further impose two trajectory-level constraints. First, ``temporal cue separation'' places the evidence required for a query in distant parts of the scenario, inserting intervening rooms and viewpoint changes so that short-window recognition is insufficient. Second, ``spatial mapping'' constructs trajectories that cannot be solved using local navigation cues alone, requiring the model to integrate observations across the traversal into a globally consistent spatial topology. These constraints enable tasks such as Memory--Action and Global Counting, which require integrating observations across distant points in the traversal, as well as Map Direction, Map Scale, and Simulation, which require constructing a coherent and globally consistent spatial representation of the environment.

From this fully specified scene and traversal, the ground-truth answers for both extractive and abstractive queries are deterministically derived.

\smallskip\smallskip\noindent
\textbf{Step 3: Rendering and Video Recording}

\smallskip
\noindent\underline{Rendering:}
Once the scene configuration is finalized, we render the environment using \texttt{Enscape}\footnote{\texttt{Enscape} is real-time architectural visualization engine that generates photorealistic indoor scenes with consistent lighting and material properties.} to generate photorealistic indoor scenes for egocentric video capture, as illustrated in Figure \ref{fig:pipeline}(e). This process converts the designed indoor environment, including the room layout and query-driven object placements, into a realistic and navigable 3D space where the camera can traverse the rooms and observe the placed objects along the predefined trajectory.

\smallskip
\noindent\underline{Video Recording:} After rendering the environment, we record egocentric videos by executing the predefined camera trajectories within the rendered 3D scene. The camera follows the navigation paths designed in Step 2, revealing observations according to the planned trajectory and evidence placement. This ensures that the evidence required to answer each query appears along the traversal in the intended spatial and temporal configuration. All videos are recorded at $30$ fps with a resolution of $640\times480$.

\subsection{Dataset Overview and Human Process}

\myalg{} consists of 10 egocentric navigation videos, each corresponding to a distinct scenario designed to support evaluation under both extractive and abstractive reasoning regimes. For each scenario, we define 10 tasks in Table \ref{tab:task_expansion}, comprising five extractive and five abstractive tasks. For each task, three queries are constructed, resulting in a total of $30$ queries grounded in a single video (5 tasks $\times$ 2 query types $\times$ 3 queries). In total, the dataset contains 300 queries across all 10 videos, with all 30 queries per scenario grounded in the same video due to our query-conditioned construction pipeline. Each query is associated with a human-annotated ground-truth answer. Except for counting-related tasks (\emph{i.e.}, Global Counting and its extractive counterpart Object Counting), other queries are formulated as multiple-choice questions (MCQs), although they can also be evaluated in an open-ended generation setting in Section \ref{sec:non-mcq}. Table \ref{tab:dataset_composition} provides additional dataset statistics, including the average number of rooms and objects per scenario and video duration (in seconds) across the 10 scenarios.

All scenarios and queries are designed through a human-driven process. We recruited graduate-level students who have each published at least one paper at a top-tier computer vision conference to design the scenarios and construct the queries. Based on the designed queries, we generate the corresponding environments and videos following our pipeline, with each video requiring approximately 2--3 weeks to produce. After generation, the annotators perform a cross-verification process to ensure that the answer to every query can be correctly derived from the video, jointly reviewing the videos to confirm that each query is unambiguously grounded in the visual evidence. Further details of the human-driven process are provided in the supplementary material.

\begin{table}[t]
    \centering
    \scriptsize
    \begin{tabularx}{\textwidth}{
        >{\hsize=0.9\hsize\centering\arraybackslash}X 
        >{\hsize=0.9\hsize\centering\arraybackslash}X 
        >{\hsize=0.8\hsize\centering\arraybackslash}X 
        >{\hsize=0.9\hsize\centering\arraybackslash}X 
        >{\hsize=1.5\hsize\centering\arraybackslash}X}
        \toprule
        \textbf{Resolution} & \textbf{\#Query} & \textbf{\#Rooms} & \textbf{\#Objects} & \textbf{Video Duration(s)}\\
        \midrule
        $640 \times 480$ & 30 & 6.7 \tiny{$\pm$ 1.25} & 73.5 \tiny{$\pm$ 14.27} & 36.81 \tiny{$\pm$ 6.05} \\
        \bottomrule
    \end{tabularx}
    \caption{Dataset statistics of \myalg{} averaged over 10 scenarios.}
    \label{tab:dataset_composition}
\end{table}

\vspace*{-0.1cm}
\section{Evaluation}
\label{sec:eval}
\vspace*{-0.1cm}

We use VAEX-Bench to quantify how current MLLMs differ in extractive vs. abstractive spatiotemporal reasoning from egocentric videos (see Section \ref{sec:main_exp}). Our main experiment benchmarks 14 SOTA proprietary or open-source MLLMs on five abstractive tasks and their extractive counterparts under a unified protocol, and further includes an MCQ-to-free-form generation comparison (see Section \ref{sec:non-mcq}) and targeted diagnostic analyses that attribute failures to temporal, spatial, and perceptual bottlenecks (see Section \ref{sec:bottleneck}).

To ensure a fair comparison, we standardize the evaluation pipeline across all MLLMs whenever the interface allows explicit control over the inputs. Similar to recent works \citep{akbari2021vatt, wasim2023vita, chen2026learning}, we uniformly sample 32 frames from each video for all models that accept video input, except for the \texttt{Gemini} family, whose API does not expose controllable frame sampling \citep{yang2025thinking}. We also enforce a unified prompt format (see the supplementary material) by prepending a fixed instruction header and constraining the output format (\emph{e.g.}, selecting a single option for MCQ).

\subsection{Main Result: Abstractive vs. Extractive Reasoning} \label{sec:main_exp}
\newcolumntype{C}{>{\centering\arraybackslash}X}

\begin{table}[t]
    \centering
    \scriptsize
    \raggedright \tiny {Mem.: Memory--Action, MDir.: Map Direction, MSca.: Map Scale, Sim.: Simulation, Glob.: Global Counting, Order: Appear Order, Dir.: Relative Direction, Dist.: Relative Distance, Route: Route plan, Obj.: Object Counting}
    \vspace{2pt}
    \begin{tabularx}{\textwidth}{l CCCCCCCCCCCC}
        \toprule
        & \multicolumn{6}{c}{\textbf{Abstractive}} & \multicolumn{6}{c}{\textbf{Extractive}} \\
        \cmidrule(lr){2-7} \cmidrule(lr){8-13}
        \textbf{Models} & Mem. & MDir. & MSca. & Sim. & Glob. & \textbf{Avg.} & Order & Dir. & Dist. & Route & Obj. & \textbf{Avg.} \\
        \hline
        \rowcolor{blue!5}
        \textit{Reference Baselines} & & & & & & & & & & & & \\
        Random & 30.7 & 22.0 & 18.0 & 35.3 & N/A & 26.5 & 15.3 & 22.0 & 42.0 & 20.0 & N/A & 24.8 \\
        Human-Level & 89.3 & 83.3 & 60.0 & 93.3 & 82.7 & 81.7 & 93.3 & 88.0 & 77.3 & 88.7 & 92.7 & 88.0 \\
        \hline
        \rowcolor{blue!5}
        \textit{Proprietary Models} & & & & & & & & & & & & \\
        GPT-5.2~\citep{singh2025openai} & 38.0 & 26.0 & 34.0 & 29.3 & 23.3 & 30.1 & 73.3 & 26.0 & 56.0 & 18.7 & 48.7 & 44.5 \\
        Gemini-3 Flash~\citep{comanici2025gemini} & 60.7 & 34.0 & 24.0 & 51.3 & 31.3 & 40.3 & 93.3 & 40.0 & 62.7 & 16.0 & 38.0 & 50.0 \\
        Gemini-3 Pro~\citep{comanici2025gemini} & 52.0 & 22.7 & 22.0 & 31.3 & 20.7 & 29.7 & 96.0 & 44.7 & 66.7 & 26.0 & 44.7 & 55.6 \\
        Claude 4.5 Haiku~\citep{anthropic2025claude4} & 19.3 & 21.3 & 11.3 & 22.0 & 2.7 & 15.3 & 34.0 & 30.0 & 26.7 & 30.0 & 40.0 & 32.1 \\
        Claude 4.5 Sonnet~\citep{anthropic2025claude4} & 30.0 & 19.3 & 19.3 & 40.7 & 11.3 & 24.1 & 72.7 & 24.7 & 28.0 & 13.3 & 30.7 & 33.9 \\
        \hline
        \rowcolor{blue!5}
        \textit{Open-Source Models} & & & & & & & & & & & & \\
        Qwen3-VL-2B~\citep{bai2025qwen3} & 26.0 & 18.7 & 19.3 & 26.0 & 6.7 & 19.3 & 82.0 & 29.3 & 42.7 & 22.7 & 40.0 & 43.3 \\
        Qwen3-VL-8B~\citep{bai2025qwen3} & 34.0 & 22.0 & 15.3 & 34.0 & 17.3 & 24.5 & 80.0 & 33.3 & 54.0 & 25.3 & 34.0 & 45.3 \\
        Qwen3-VL-32B~\citep{bai2025qwen3} & 40.0 & 26.0 & 23.3 & 42.7 & 17.3 & 29.9 & 90.0 & 26.0 & 53.3 & 6.7 & 51.3 & 45.5 \\
        Qwen3-VL-235B$^{\dagger}$~\citep{bai2025qwen3} & 43.3 & 16.7 & 13.3 & 46.7 & 13.3 & 26.7 & 93.3 & 38.0 & 54.0 & 20.0 & 43.3 & 49.7 \\
        InternVL3.5-2B~\citep{wang2025internvl3} & 28.0 & 22.7 & 20.0 & 21.3 & 14.0 & 21.2 & 64.0 & 30.7 & 36.7 & 20.0 & 30.7 & 36.4 \\
        InternVL3.5-8B~\citep{wang2025internvl3} & 36.7 & 24.0 & 21.3 & 26.0 & 16.0 & 24.8 & 73.3 & 28.7 & 36.0 & 20.7 & 32.0 & 38.1 \\  
        Mistral-3.2-24B~\citep{liu2026ministral} & 28.7 & 16.7 & 10.0 & 18.7 & 14.0 & 17.6 & 28.7 & 28.0 & 40.0 & 17.3 & 40.7 & 30.9\\
        Gemma-3-27B~\citep{Kamath2025Gemma3T} & 32.0 & 24.0 & 23.3 & 28.7 & 10.7 & 23.7 & 55.3 & 33.3 & 28.0 & 28.7 & 49.3 & 38.9 \\
        GLM-4.6V-106B~\citep{hong2025glm} & 35.3 & 18.0 & 20.7 & 37.3 & 6.0 &  23.5 & 85.3 & 29.3 & 58.4 & 10.0 & 52.0 & 46.9 \\
        \bottomrule
    \end{tabularx}
    \caption{Performance on \myalg{} over 14 SOTA MLLMs compared with Random and Human baselines. $\dagger$ denotes a MoE model with 22B activated parameters.}
    \label{tab:model_comparison}
\end{table}

Table \ref{tab:model_comparison} reports the performance on \myalg{}, highlighting a clear gap between extractive tasks grounded in explicit visual evidence and abstractive tasks requiring integration of observations across time and space.

\smallskip\smallskip\noindent
\textbf{Evaluation Setup.} 
We evaluate in a zero-shot setting using each MLLM's default interface. Following prior work in MLLM evaluation \citep{yang2025thinking}, we adopt a fixed decoding configuration across MLLMs with temperature $0.7$, top-$p$ $=1.0$, and top-$k$ $=40$. We evaluate stochastic decoding stability by reporting Accuracy@k, defined as the average accuracy across $k$ independent generations per query. In our experiments, we set $k=5$ and report Accuracy@5 for both MCQ and numerical-answer items. To provide reference points, we include two baselines: ``Random'' is the chance-level accuracy obtained by selecting an answer without video input for MCQ queries \citep{yang2025thinking} (N/A indicates Counting questions where random chance is not applicable), while ``Human'' reports the average accuracy of five human annotators who first watch the corresponding egocentric video. 

\smallskip\smallskip\noindent\textbf{Performance of Humans.}
We include the Human reference to validate task difficulty and confirm that the questions are solvable from the provided egocentric evidence. While its performance is high and relatively uniform on extractive tasks (88.0\% on average), it drops to 81.7\% on abstractive tasks, suggesting that abstractive reasoning imposes additional cognitive demands beyond recognizing visible cues. Within the abstractive setting, performance is particularly low on {Map Scale} (MSca.), reflecting the intrinsic difficulty of estimating relative metric distances even when one can mentally reconstruct an approximate floor-plan from egocentric observations. Interestingly, even in the extractive setting, distance questions (Dist.) remain comparatively harder, highlighting a shared bottleneck in metric intuition from visual observations.

\smallskip\smallskip\noindent\textbf{Performance of Proprietary MLLMs.}
The proprietary MLLMs outperform the Random baseline, showing fairly reasonable performance on extractive tasks. However, their performance drops substantially on abstractive tasks. Overall, the \texttt{Gemini} family generally outperforms the \texttt{Claude} family across both settings. Interestingly, the performance ranking also shifts between the two tasks: \texttt{Gemini-3 Pro} leads on extractive tasks, yet \texttt{Gemini-3 Flash} performs substantially better on abstractive tasks, indicating that improvements in short-horizon recognition are unlikely to translate to abstractive spatiotemporal reasoning.

\smallskip\smallskip\noindent\textbf{Performance of Open-Source MLLMs.}
Like proprietary MLLMs, open-source models perform poorly on abstractive tasks relative to extractive ones. Even when comparing the most powerful open-source MLLMs (\emph{i.e.}, \texttt{Qwen3-VL-32} \texttt{/235B}) to proprietary MLLMs, the performance gap is relatively large in the abstractive setting, while it is relatively small on extractive tasks. That is, for extractive tasks, several open-source MLLMs are competitive with proprietary counterparts, indicating that performance limitations mainly arise in abstractive spatiotemporal reasoning rather than extractive perception. Moreover, within open-source MLLMs, scaling does not yield monotonic gains on abstractive tasks. A particularly consistent weakness across both proprietary and open-source MLLMs is {Global Counting} (Glob.), where models remain far from human-level performance, pointing to persistent failures in entity persistence and duplicate-aware aggregation under partial observability, as detailed in Section \ref{sec:bottleneck}.

\subsection{From MCQ to Free-form Generation} \label{sec:non-mcq}
\begin{figure}[t]
    \centering
    \begin{minipage}{0.6\textwidth}
        \centering
        \scriptsize
        \raggedright \tiny ~~ {Mem.: Memory--Action, MDir.: Map Direction, MSca.: Map Scale}
        \begin{tabularx}{\textwidth}{X cccc}
            \toprule
            Models & Mem. & MDir. & MSca. & Avg. \\
            \midrule
            GPT-5.2         & 35.3  \tiny{\textcolor{red}{($\downarrow$2.7)}}& 15.3 \tiny{\textcolor{red}{($\downarrow$10.7)}} & 12.0 \tiny{\textcolor{red}{($\downarrow$22.0)}} &  20.9 \tiny{\textcolor{red}{($\downarrow$11.8)}} \\
            Gemini-3-Flash    & 49.3 \tiny{\textcolor{red}{($\downarrow$11.3)}} & 13.3 \tiny{\textcolor{red}{($\downarrow$20.7)}} & 11.3 \tiny{\textcolor{red}{($\downarrow$12.7)}} &  24.7 \tiny{\textcolor{red}{($\downarrow$14.9)}}  \\
            Gemini-3-Pro    & 52.0 \tiny{\textcolor{red}{($\downarrow$0.0)}} & 18.7 \tiny{\textcolor{red}{($\downarrow$4.0)}}& 10.0 \tiny{\textcolor{red}{($\downarrow$12.0)}}&  26.9 \tiny{\textcolor{red}{($\downarrow$5.3)}}  \\
            Qwen3-VL-32B    & 38.0 \tiny{\textcolor{red}{($\downarrow$2.0)}} & 19.3 \tiny{\textcolor{red}{($\downarrow$6.7)}}& 11.3 \tiny{\textcolor{red}{($\downarrow$12.0)}} &  22.9 \tiny{\textcolor{red}{($\downarrow$6.9)}} \\
            InternVL3.5-8B    & 36.0 \tiny{\textcolor{red}{($\downarrow$0.7)}}& 18.7 \tiny{\textcolor{red}{($\downarrow$5.3)}}& 13.3 \tiny{\textcolor{red}{($\downarrow$8.0)}}  &  22.4 \tiny{\textcolor{red}{($\downarrow$4.7)}} \\
            Gemma-3-27B    & 31.3 \tiny{\textcolor{red}{($\downarrow$0.7)}} & 16.0 \tiny{\textcolor{red}{($\downarrow$8.0)}} & 11.3 \tiny{\textcolor{red}{($\downarrow$12.0)}} &  19.6 \tiny{\textcolor{red}{($\downarrow$6.9)}} \\
            \bottomrule
        \end{tabularx}
        \captionof{table}{{Performance on \myalg{} in free-form generation, where the MCQ-format questions of the abstractive tasks are converted into open-ended queries without providing answer choices.}}
        \label{tab:generation_judge}
    \end{minipage}
    \hfill
    \begin{minipage}{0.37\textwidth}
        \centering
        \vspace*{-0.3cm}
        \includegraphics[width=\textwidth]{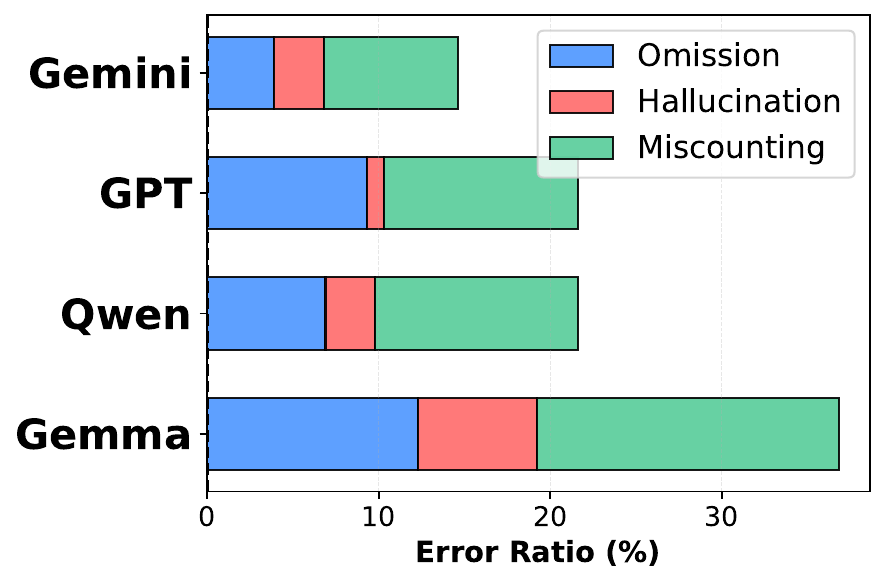}
        \vspace*{-0.5cm}
        \captionof{figure}{Error breakdown for the Global Counting task.}
        \label{fig:cumul}
    \end{minipage}
\end{figure}

We additionally evaluate MLLMs in a free-form generation setting, where predictions are produced as text rather than selected from multiple-choice options. Except for the Counting task, the other four abstractive tasks are converted from MCQ to free-form. For comparison, we select the best-performing MLLMs under the MCQ setting in Table \ref{tab:model_comparison}, including the 3 best proprietary and the 3 best open-source MLLMs. 
We evaluate the correctness of MLLM-generated responses using an LLM-as-a-judge framework, with the ground-truth answer provided as reference, since the Memory--Action task requires open-ended responses that may vary in wording while conveying the same underlying action sequence. In contrast, for the other tasks, we parse the generated output into a structured answer and apply exact matching against the ground-truth. 

As summarized in Table \ref{tab:generation_judge}, replacing MCQ with free-form generation yields a clear performance drop, since MLLMs can no longer benefit from option elimination or chance selection and must explicitly produce the correct content in the expected form. This suggests that MCQ-based evaluation may partially overestimate model capabilities and that free-form generation provides a more stringent test of abstractive spatiotemporal reasoning.

\subsection{Detailed Analysis \& Discussion}\label{sec:bottleneck}

\begin{figure}[t]
  \centering
  
  \begin{subfigure}[b]{0.48\textwidth}
    \centering
    \includegraphics[width=\textwidth]{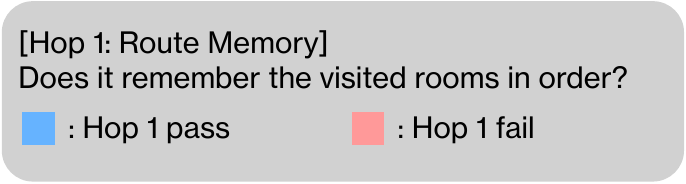} 
  \end{subfigure}
  \begin{subfigure}[b]{0.48\textwidth}
    \centering
    \includegraphics[width=\textwidth]{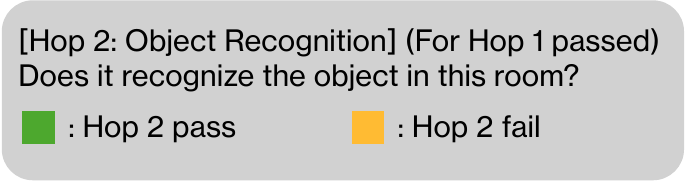}
  \end{subfigure} 
  
  \begin{subfigure}[b]{0.24\textwidth}
    \centering
    \includegraphics[width=\textwidth]{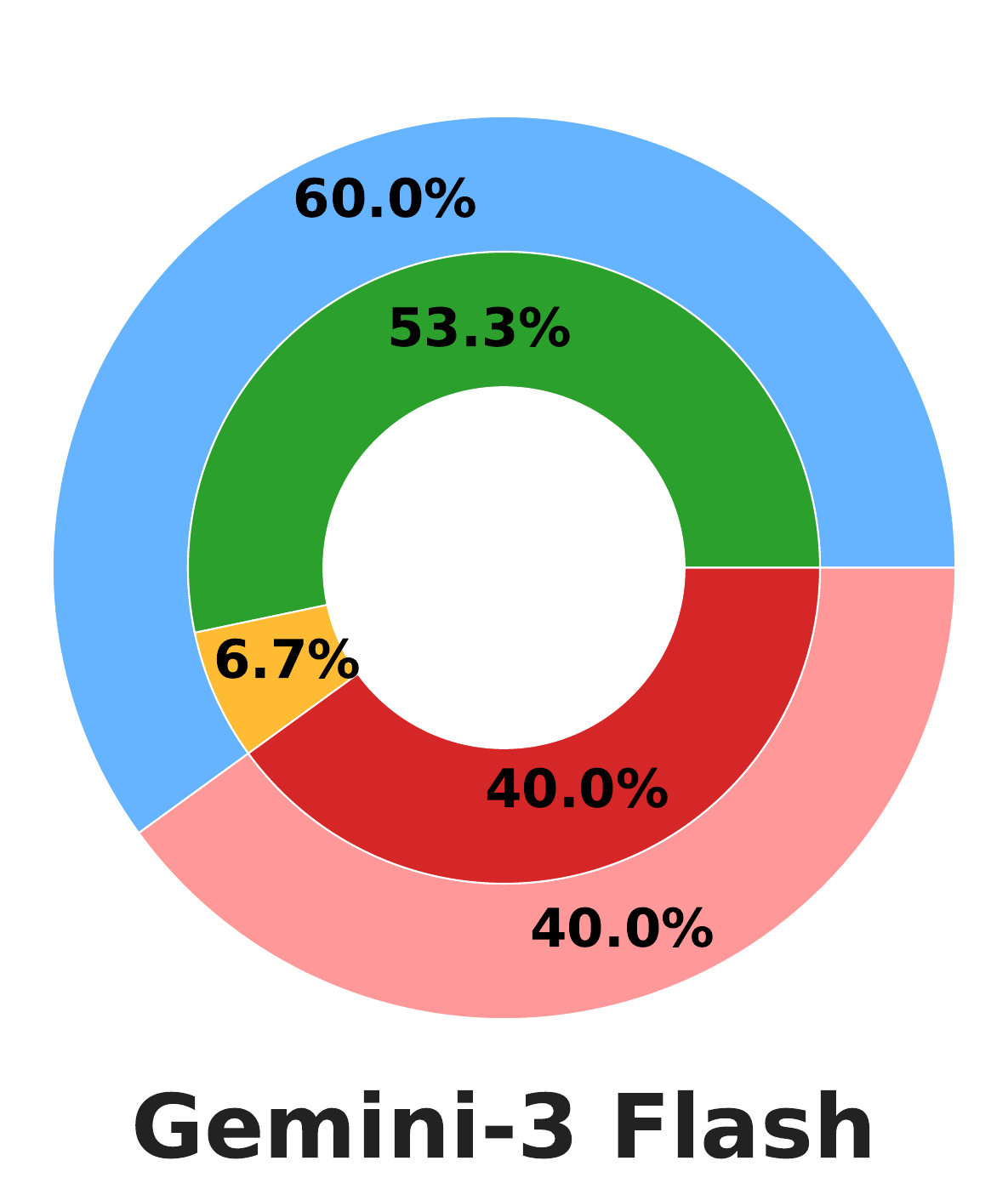}
  \end{subfigure}
  \hfill
  \begin{subfigure}[b]{0.24\textwidth}
    \centering
    \includegraphics[width=\textwidth]{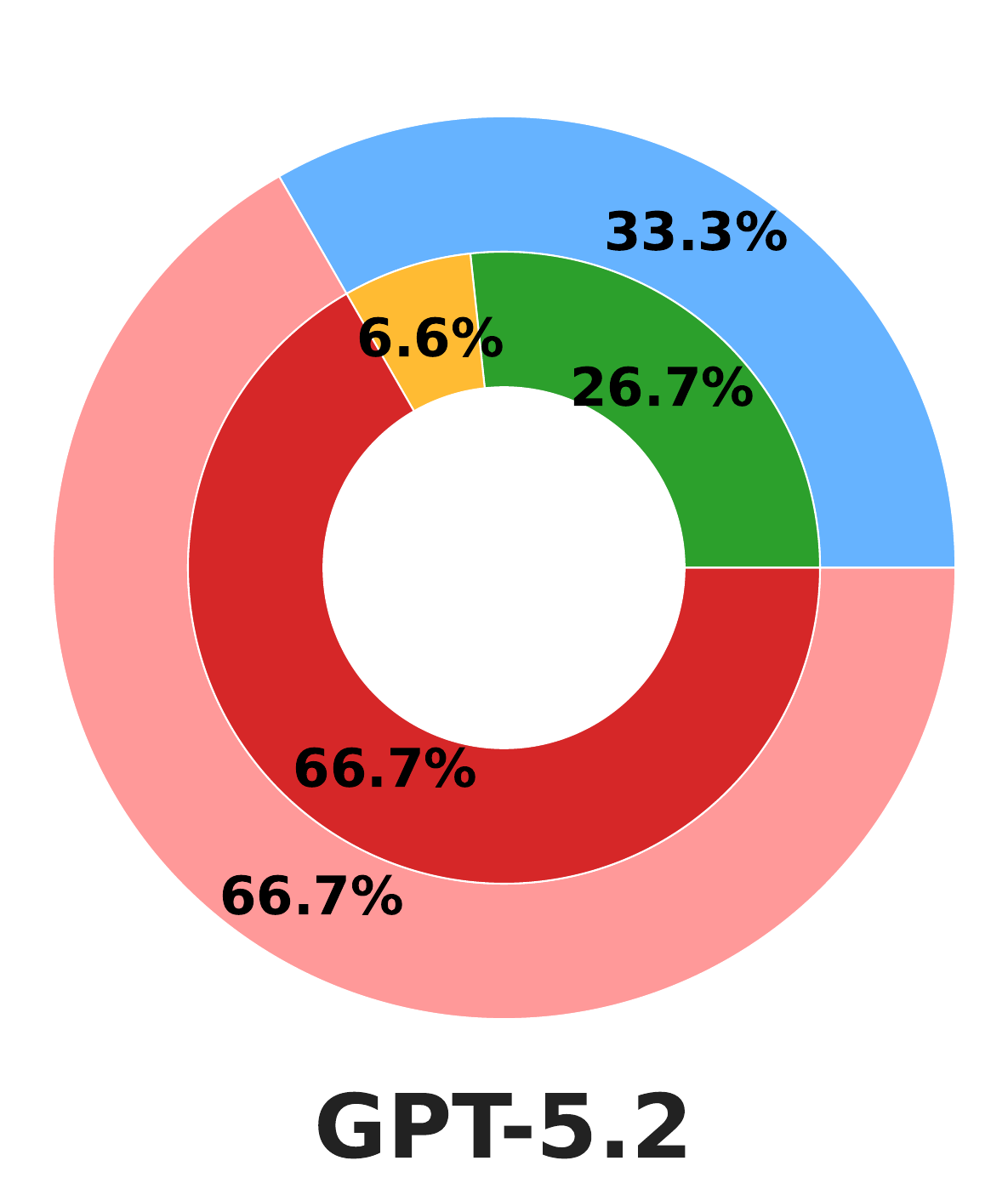}
  \end{subfigure}
  \hfill
  \begin{subfigure}[b]{0.24\textwidth}
    \centering
    \includegraphics[width=\textwidth]{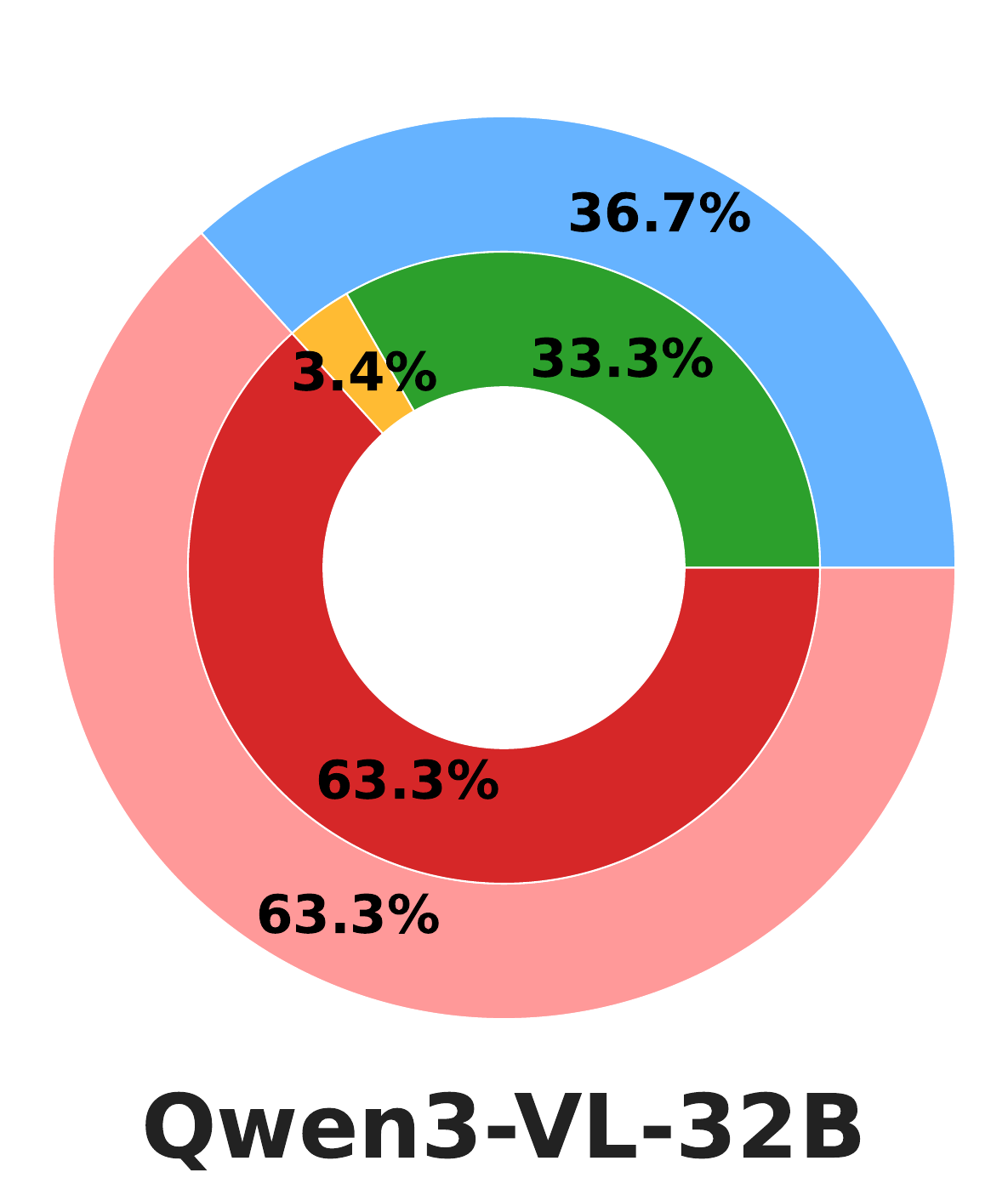}
  \end{subfigure}
  \begin{subfigure}[b]{0.24\textwidth}
    \centering
    \includegraphics[width=\textwidth]{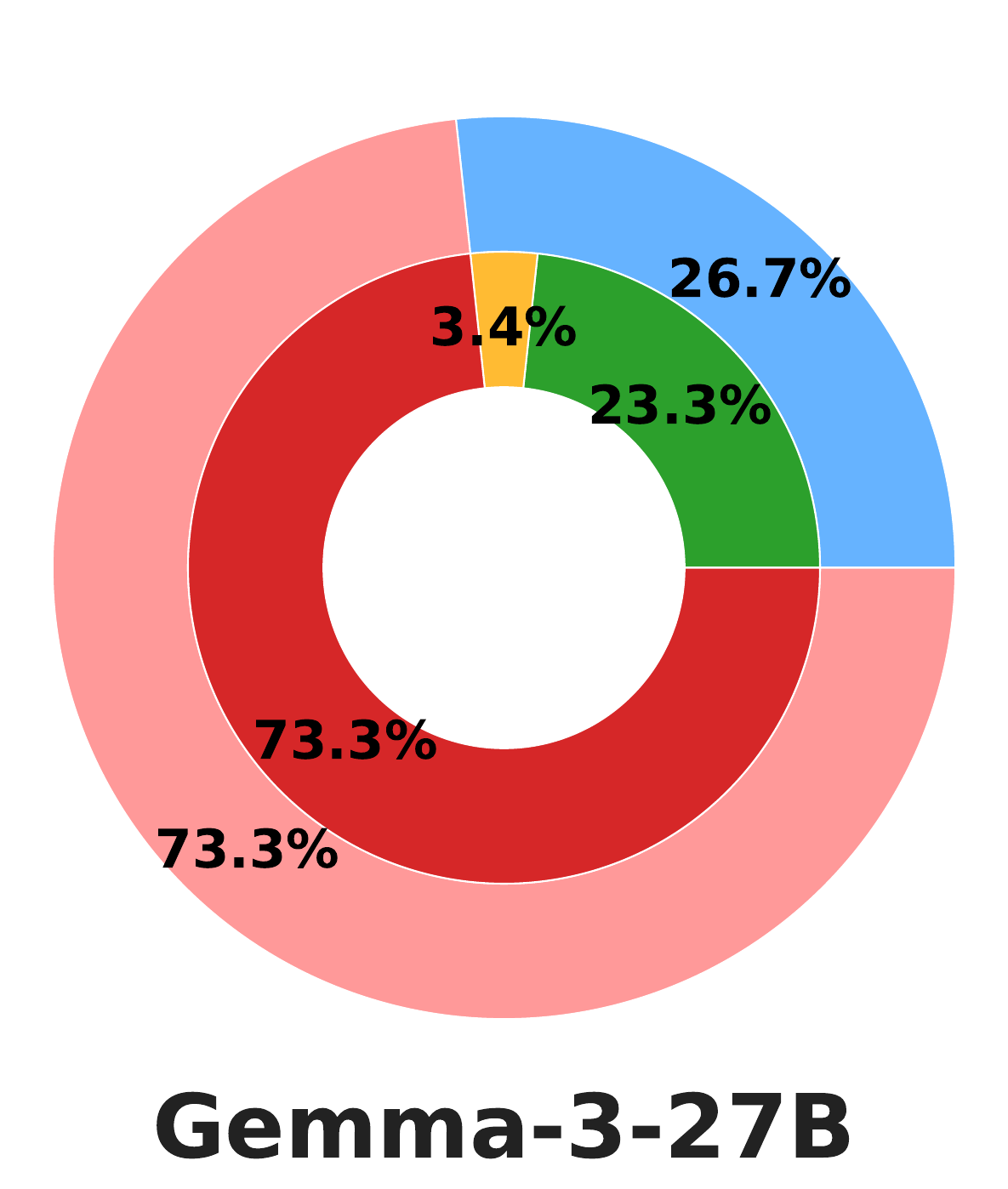}
  \end{subfigure}
  \caption{{Temporal bottleneck in multi-hop sub-tasks.} The outer circle represents the pass/fail ratio for Hop 1 (Blue: Pass, Pink: Fail). The inner ring shows the conditional pass rate for Hop 2 among the cases that passed Hop 1 (Green: Pass, Yellow: Fail).}
  \label{fig:temporal}
\end{figure}

\begin{figure}[t]
    \centering
    \includegraphics[width=\textwidth]{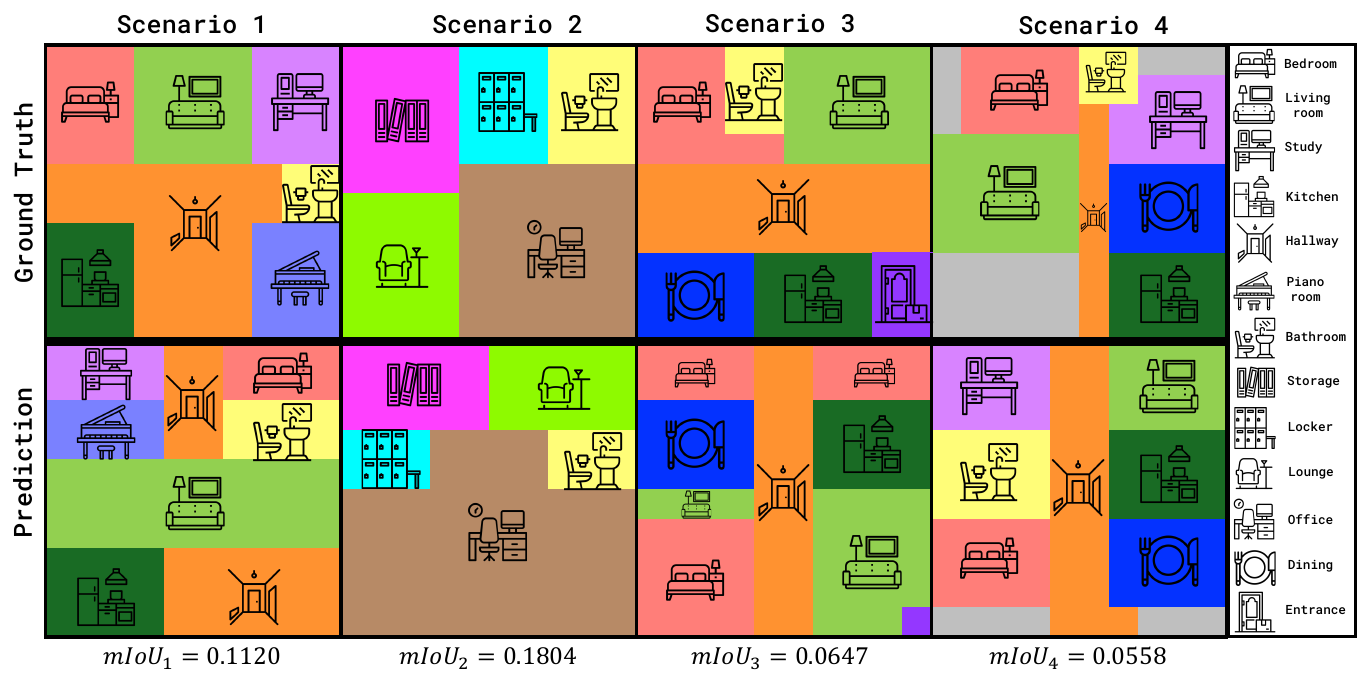}
    \caption{{Visualization of floor-plan prediction.} The upper and lower halves, separated by a bold dark line, represent the Ground Truth and Prediction, respectively. Colored icons denote functional areas, while grey regions indicate ``None''.}
    \label{fig:floor_plan}
\end{figure}

In this section, we move beyond aggregate accuracy to diagnose why current models fail at abstractive spatiotemporal reasoning. To pinpoint the underlying causes, we decompose each abstractive task into finer-grained sub-tasks and analyze performance across these components. Complementing this quantitative breakdown, we conduct qualitative analyses and visualizations (\emph{e.g.}, floor-plan prediction) to make model behavior interpretable. Across these analyses, we identify three recurring bottlenecks:  {``Perceptual''} (Figure \ref{fig:cumul}), {``Temporal''} (Figure \ref{fig:temporal}), and {``Spatial''} (Figure \ref{fig:floor_plan}). For this analysis, we select four high-performing MLLMs: two from the proprietary models (\texttt{Gemini-3 Flash} and \texttt{GPT-5.2}) and another two from the open-source models (\texttt{Qwen3-VL-32B} and \texttt{Gemma-3-27B}).

\smallskip\smallskip\noindent\textbf{Perceptual Bottleneck.} We analyze failures on {Global Counting}, an abstractive reasoning task that strongly depends on reliable perception across space and time, requiring models to detect and track object instances before aggregating them into global counts. To obtain a granular view of model failures, we decompose errors into three mutually exclusive categories: \emph{Omission}, where a present object is counted as absent; \emph{Hallucination}, where a non-existent object is counted as present; and \emph{Miscounting}, where the object exists but its count is incorrect. Such errors can arise either at the room level during local perception or during the aggregation process when combining counts across rooms to form the final scenario-level estimate. Figure~\ref{fig:cumul} illustrates the distribution of these error categories across the failure cases.

Across models, Miscounting accounts for the largest share of errors, indicating that even when objects are perceived, models often produce incorrect counts within rooms or during cross-room aggregation. Omission also contributes substantially, suggesting that objects are frequently missed due to occlusion or limited viewpoints. In contrast, Hallucination errors are relatively rare, implying that the main difficulty lies in accurately maintaining and aggregating object counts across the video rather than detecting objects that do not exist.

\smallskip\smallskip\noindent\textbf{Temporal Bottleneck.} To analyze temporal failures in long-horizon reasoning, we examine the Memory--Action task, which requires models to recall the sequence of visited rooms along the trajectory and use that temporal memory to identify the correct action associated with a queried location. To disentangle different sources of failure, we perform a fine-grained error analysis along two hierarchical components: \emph{Global Temporal Memory} (Hop 1: tracking the sequence of visited rooms) and \emph{Local Temporal Memory} (Hop 2: recalling the objects observed in the corresponding room for action reasoning). Notably, failing Hop 1 necessarily implies failing Hop 2, since incorrect room identification prevents verifying the correct in-room objects. Accordingly, as illustrated in Figure \ref{fig:temporal}, each prediction falls into one of three interpretable outcomes: \emph{Success} (Hop 1 $\checkmark$, Hop 2 $\checkmark$) indicates that the model correctly reconstructs the visited-room sequence and recalls the relevant objects within the queried room, \emph{Local Memory Error} (Hop 1 $\checkmark$, Hop 2 $\times$) occurs when the model identifies the correct room in the trajectory but fails to recall the objects observed there, and \emph{Global Memory Error} (Hop 1 $\times$, Hop 2 $\times$) arises when the model fails to recover the room visitation order, which consequently prevents correct object recall.

As shown in Figure \ref{fig:temporal}, the dominant failure mode across models is Global Memory Error, indicating that many predictions already fail at Hop 1 by incorrectly reconstructing the visited-room sequence. Even when Hop 1 is successful, the conditional pass rate of Hop 2 remains relatively moderate, suggesting additional difficulty in recalling the objects associated with the queried room. Overall, these results indicate that maintaining the correct temporal order of visited locations is the primary bottleneck, while local object recall becomes a secondary source of error once the global trajectory memory is established.

\smallskip\smallskip\noindent\textbf{Spatial Bottleneck.} Remaining tasks (Map Direction, Map Scale, and Simulation) fundamentally require a coherent global spatial representation of the environment. Models must reason about relative room orientation, maintain consistent metric proportions, and simulate viewpoint transformations within a unified layout. To diagnose these spatial bottlenecks, we evaluate the MLLM's capacity to explicitly reconstruct a global cognitive map via grid-based floor-plan prediction, as shown in Figure \ref{fig:floor_plan}. Concretely, we provide the MLLM with a $10\times10$ empty grid, a fixed label set consisting of the candidate room types (including ``None'') appearing in the scenario, and the camera start location on the grid as an anchor point. The model is instructed to output the floor plan as $10$ lines, each containing $10$ cell labels (one per grid position) per row.  We then manually render the predicted grid by coloring each cell according to the generated labels for visualization and analysis. 

To quantitatively evaluate spatial reconstruction quality, we compute mean Intersection-over-Union (mIoU) between the predicted and ground-truth room segments on the grid. We use \texttt{Gemini-3 Flash} for this analysis due to its strong overall performance; results for other models, along with detailed metric definitions, prompts, and formatting constraints, are in the supplementary material.

Accurately populating this 2D grid requires not only tracking topological adjacency but also maintaining consistent metric proportions and absolute spatial alignment across the trajectory. As illustrated in Figure \ref{fig:floor_plan}, we observe substantial mismatches between the predicted layouts and the ground-truth floor plans. In particular, models frequently produce distorted room boundaries and inconsistent grid allocations, indicating difficulty in maintaining metric scale. We also observe rotational misalignments and incorrect room adjacencies, suggesting that models struggle to anchor a consistent global coordinate frame. Finally, the reconstructed layouts are often fragmented and spatially incoherent, preventing models from reliably simulating viewpoint transformations. Together, these observations indicate that current MLLMs struggle to construct a coherent global spatial representation of the environment.

\section{Conclusion}

We present \myalg{}, a scenario-driven benchmark for evaluating abstractive and extractive spatiotemporal reasoning in egocentric videos, moving beyond evidence-present extraction toward long-horizon memory, allocentric map reasoning, and global evidence aggregation. Unlike prior benchmarks built on fixed captured footage, \myalg{} is manually constructed with controlled scenario design and spatially consistent QA pairs, enabling precise diagnosis of failure modes. Our experiments reveal persistent bottlenecks in global spatial memory and structured evidence integration across state-of-the-art MLLMs. Future work will focus on automating the pipeline to scale benchmark generation while preserving its controllability.

\noindent \textbf{Limitations.}
A primary limitation of \myalg{} is its modest scale. Each scenario is manually crafted end to end, including scenario specification, floor-plan design, object modeling and placement, rendering and capture, and human QA construction, so the number of videos is necessarily limited. We view this as a deliberate trade-off in favor of rigor and diagnostic control. The pipeline enforces strict spatial consistency, an unambiguous answer, and targeted scenario factors that are difficult to guarantee in large-scale captured footage. As a result, \myalg{} is intended less as a large corpus and more as a high-precision testbed for isolating failure modes, long-horizon spatiotemporal reasoning, with scaling through partial automation as a clear direction for future work.

\bibliographystyle{assets/plainnat}
\bibliography{colm2026_conference}

@article{bai2025qwen3,
  title={Qwen3-vl technical report},
  author={Bai, Shuai and Cai, Yuxuan and Chen, Ruizhe and Chen, Keqin and Chen, Xionghui and Cheng, Zesen and Deng, Lianghao and Ding, Wei and Gao, Chang and Ge, Chunjiang and others},
  journal={arXiv preprint arXiv:2511.21631},
  year={2025}
}

@article{wang2025internvl3,
  title={Internvl3. 5: Advancing open-source multimodal models in versatility, reasoning, and efficiency},
  author={Wang, Weiyun and Gao, Zhangwei and Gu, Lixin and Pu, Hengjun and Cui, Long and Wei, Xingguang and Liu, Zhaoyang and Jing, Linglin and Ye, Shenglong and Shao, Jie and others},
  journal={arXiv preprint arXiv:2508.18265},
  year={2025}
}

@article{singh2025openai,
  title={Openai gpt-5 system card},
  author={Singh, Aaditya and Fry, Adam and Perelman, Adam and Tart, Adam and Ganesh, Adi and El-Kishky, Ahmed and McLaughlin, Aidan and Low, Aiden and Ostrow, AJ and Ananthram, Akhila and others},
  journal={arXiv preprint arXiv:2601.03267},
  year={2025}
}

@article{Kamath2025Gemma3T,
  title={Gemma 3 Technical Report},
  author={Aishwarya Kamath, Johan Ferret, Shreya Pathak and others},
  journal={arXiv preprint arXiv:2601.03267},
  year={2025}
}

@article{comanici2025gemini,
  title={Gemini 2.5: Pushing the frontier with advanced reasoning, multimodality, long context, and next generation agentic capabilities},
  author={Comanici, Gheorghe and Bieber, Eric and Schaekermann, Mike and Pasupat, Ice and Sachdeva, Noveen and Dhillon, Inderjit and Blistein, Marcel and Ram, Ori and Zhang, Dan and Rosen, Evan and others},
  journal={arXiv preprint arXiv:2507.06261},
  year={2025}
}

@article{hong2025glm,
  title={Glm-4.5 v and glm-4.1 v-thinking: Towards versatile multimodal reasoning with scalable reinforcement learning},
  author={Hong, Wenyi and Yu, Wenmeng and Gu, Xiaotao and Wang, Guo and Gan, Guobing and Tang, Haomiao and Cheng, Jiale and Qi, Ji and Ji, Junhui and Pan, Lihang and others},
  journal={arXiv preprint arXiv:2507.01006},
  year={2025}
}

@article{cheng2025v,
  title={V-star: Benchmarking video-llms on video spatio-temporal reasoning},
  author={Cheng, Zixu and Hu, Jian and Liu, Ziquan and Si, Chenyang and Li, Wei and Gong, Shaogang},
  journal={arXiv preprint arXiv:2503.11495},
  year={2025}
}

@article{ning2025video,
  title={Video-bench: A comprehensive benchmark and toolkit for evaluating video-based large language models},
  author={Ning, Munan and Zhu, Bin and Xie, Yujia and Lin, Bin and Cui, Jiaxi and Yuan, Lu and Chen, Dongdong and Yuan, Li},
  journal={Computational Visual Media},
  year={2025}
}

@inproceedings{rodin2025easg,
  title={EASG-Bench: Video Q\&A Benchmark with Egocentric Action Scene Graphs},
  author={Rodin, Ivan and Wu, Tz-Ying and Min, Kyle and Sridhar, Sharath Nittur and Furnari, Antonino and Tripathi, Subarna and Farinella, Giovanni Maria},
  booktitle={ICCV},
  year={2025}
}

@inproceedings{dai2017scannet,
  title={Scannet: Richly-annotated 3d reconstructions of indoor scenes},
  author={Dai, Angela and Chang, Angel X and Savva, Manolis and Halber, Maciej and Funkhouser, Thomas and Nie{\ss}ner, Matthias},
  booktitle={CVPR},
  year={2017}
}

@inproceedings{yeshwanth2023scannet++,
  title={Scannet++: A high-fidelity dataset of 3d indoor scenes},
  author={Yeshwanth, Chandan and Liu, Yueh-Cheng and Nie{\ss}ner, Matthias and Dai, Angela},
  booktitle={ICCV},
  year={2023}
}

@article{baruch2021arkitscenes,
  title={Arkitscenes: A diverse real-world dataset for 3d indoor scene understanding using mobile rgb-d data},
  author={Baruch, Gilad and Chen, Zhuoyuan and Dehghan, Afshin and Dimry, Tal and Feigin, Yuri and Fu, Peter and Gebauer, Thomas and Joffe, Brandon and Kurz, Daniel and Schwartz, Arik and others},
  journal={arXiv preprint arXiv:2111.08897},
  year={2021}
}

@inproceedings{wang2025lvbench,
  title={Lvbench: An extreme long video understanding benchmark},
  author={Wang, Weihan and He, Zehai and Hong, Wenyi and Cheng, Yean and Zhang, Xiaohan and Qi, Ji and Ding, Ming and Gu, Xiaotao and Huang, Shiyu and Xu, Bin and others},
  booktitle={ICCV},
  year={2025}
}

@inproceedings{kesenvilma,
  title={ViLMA: A Zero-Shot Benchmark for Linguistic and Temporal Grounding in Video-Language Models},
  author={Kesen, Ilker and Pedrotti, Andrea and Dogan, Mustafa and Cafagna, Michele and Acikgoz, Emre Can and Parcalabescu, Letitia and Calixto, Iacer and Frank, Anette and Gatt, Albert and Erdem, Aykut and others},
  booktitle={ICLR},
  year={2024}
}

@inproceedings{zhou2025vlm4d,
  title={Vlm4d: Towards spatiotemporal awareness in vision language models},
  author={Zhou, Shijie and Vilesov, Alexander and He, Xuehai and Wan, Ziyu and Zhang, Shuwang and Nagachandra, Aditya and Chang, Di and Chen, Dongdong and Wang, Xin Eric and Kadambi, Achuta},
  booktitle={ICCV},
  year={2025}
}

@inproceedings{wang2024sok,
  title={Sok-bench: A situated video reasoning benchmark with aligned open-world knowledge},
  author={Wang, Andong and Wu, Bo and Chen, Sunli and Chen, Zhenfang and Guan, Haotian and Lee, Wei-Ning and Li, Li Erran and Gan, Chuang},
  booktitle={CVPR},
  year={2024}
}

@article{xu2024survey,
  title={A survey of resource-efficient llm and multimodal foundation models},
  author={Xu, Mengwei and Yin, Wangsong and Cai, Dongqi and Yi, Rongjie and Xu, Daliang and Wang, Qipeng and Wu, Bingyang and Zhao, Yihao and Yang, Chen and Wang, Shihe and others},
  journal={arXiv preprint arXiv:2401.08092},
  year={2024}
}

@inproceedings{sermanet2024robovqa,
  title={Robovqa: Multimodal long-horizon reasoning for robotics},
  author={Sermanet, Pierre and Ding, Tianli and Zhao, Jeffrey and Xia, Fei and Dwibedi, Debidatta and Gopalakrishnan, Keerthana and Chan, Christine and Dulac-Arnold, Gabriel and Maddineni, Sharath and Joshi, Nikhil J and others},
  booktitle={ICRA},
  year={2024}
}

@article{liu2026ministral,
  title={Ministral 3},
  author={Liu, Alexander H and Khandelwal, Kartik and Subramanian, Sandeep and Jouault, Victor and Rastogi, Abhinav and Sad{\'e}, Adrien and Jeffares, Alan and Jiang, Albert and Cahill, Alexandre and Gavaudan, Alexandre and others},
  journal={arXiv preprint arXiv:2601.08584},
  year={2026}
}

@misc{anthropic2025claude4,
  author       = {{Anthropic}},
  title        = {System Card: Claude Opus 4 \& Claude Sonnet 4},
  howpublished = {System card},
  month        = may,
  year         = {2025},
  note         = {Anthropic. Version dated May 2025.}
}

@inproceedings{yang2025thinking,
  title={Thinking in space: How multimodal large language models see, remember, and recall spaces},
  author={Yang, Jihan and Yang, Shusheng and Gupta, Anjali W and Han, Rilyn and Fei-Fei, Li and Xie, Saining},
  booktitle={CVPR},
  year={2025}
}

@article{fan2025vlm,
  title={VLM-3R: Vision-Language Models Augmented with Instruction-Aligned 3D Reconstruction},
  author={Fan, Zhiwen and Zhang, Jian and Li, Renjie and Zhang, Junge and Chen, Runjin and Hu, Hezhen and Wang, Kevin and Qu, Huaizhi and Wang, Dilin and Yan, Zhicheng and others},
  journal={arXiv preprint arXiv:2505.20279},
  year={2025}
}

@inproceedings{castro2020lifeqa,
  title={LifeQA: A real-life dataset for video question answering},
  author={Castro, Santiago and Azab, Mahmoud and Stroud, Jonathan and Noujaim, Cristina and Wang, Ruoyao and Deng, Jia and Mihalcea, Rada},
  booktitle={LREC},
  year={2020}
}

@inproceedings{tapaswi2016movieqa,
  title={Movieqa: Understanding stories in movies through question-answering},
  author={Tapaswi, Makarand and Zhu, Yukun and Stiefelhagen, Rainer and Torralba, Antonio and Urtasun, Raquel and Fidler, Sanja},
  booktitle={CVPR},
  year={2016}
}

@inproceedings{jang2017tgif,
  title={Tgif-qa: Toward spatio-temporal reasoning in visual question answering},
  author={Jang, Yunseok and Song, Yale and Yu, Youngjae and Kim, Youngjin and Kim, Gunhee},
  booktitle={CVPR},
  year={2017}
}

@inproceedings{fu2025video,
  title={Video-mme: The first-ever comprehensive evaluation benchmark of multi-modal llms in video analysis},
  author={Fu, Chaoyou and Dai, Yuhan and Luo, Yongdong and Li, Lei and Ren, Shuhuai and Zhang, Renrui and Wang, Zihan and Zhou, Chenyu and Shen, Yunhang and Zhang, Mengdan and others},
  booktitle={CVPR},
  year={2025}
}

@inproceedings{wu2024longvideobench,
  title={Longvideobench: A benchmark for long-context interleaved video-language understanding},
  author={Wu, Haoning and Li, Dongxu and Chen, Bei and Li, Junnan},
  booktitle={NeurIPS},
  year={2024}
}

@article{sugandhika2025know,
  title={Know-Show: Benchmarking Video-Language Models on Spatio-Temporal Grounded Reasoning},
  author={Sugandhika, Chinthani and Li, Chen and Rajan, Deepu and Fernando, Basura},
  journal={arXiv preprint arXiv:2512.05513},
  year={2025}
}

@article{fan2025tool,
  title={Tool-Augmented Spatiotemporal Reasoning for Streamlining Video Question Answering Task},
  author={Fan, Sunqi and Cui, Jiashuo and Guo, Meng-Hao and Yang, Shuojin},
  journal={arXiv preprint arXiv:2512.10359},
  year={2025}
}

@inproceedings{li2023discovering,
  title={Discovering spatio-temporal rationales for video question answering},
  author={Li, Yicong and Xiao, Junbin and Feng, Chun and Wang, Xiang and Chua, Tat-Seng},
  booktitle={ICCV},
  year={2023}
}

@article{gao2025abstral,
  title={AbstRaL: Augmenting LLMs' Reasoning by Reinforcing Abstract Thinking},
  author={Gao, Silin and Bosselut, Antoine and Bengio, Samy and Abbe, Emmanuel},
  journal={arXiv preprint arXiv:2506.07751},
  year={2025}
}

@article{chen2025exploring,
  title={Exploring the hidden reasoning process of large language models by misleading them},
  author={Chen, Guanyu and Wang, Peiyang and Zhang, Tianren and Chen, Feng},
  journal={arXiv preprint arXiv:2503.16401},
  year={2025}
}

@inproceedings{jiangrethinking,
  title={Rethinking LLM Reasoning: From Explicit Trajectories to Latent Representations},
  author={Jiang, Cong and Zhang, Xiaofeng and Zhang, Zheng and Zhu, Fangzhi and Chen, XiaoWei and Zhu, Junxiong},
  booktitle={ICLR},
  year={2026}
}

@article{hao2024training,
  title={Training large language models to reason in a continuous latent space},
  author={Hao, Shibo and Sukhbaatar, Sainbayar and Su, DiJia and Li, Xian and Hu, Zhiting and Weston, Jason and Tian, Yuandong},
  journal={arXiv preprint arXiv:2412.06769},
  year={2024}
}

@inproceedings{lim2025visescape,
  title={VisEscape: A Benchmark for Evaluating Exploration-driven Decision-making in Virtual Escape Rooms},
  author={Lim, Seungwon and Kim, Sungwoong and Yu, Jihwan and Lee, Sungjae and Chung, Jiwan and Yu, Youngjae},
  booktitle={EMNLP},
  year={2025}
}

@inproceedings{yang2018hotpotqa,
  title={HotpotQA: A dataset for diverse, explainable multi-hop question answering},
  author={Yang, Zhilin and Qi, Peng and Zhang, Saizheng and Bengio, Yoshua and Cohen, William and Salakhutdinov, Ruslan and Manning, Christopher D},
  booktitle={EMNLP},
  year={2018}
}

@inproceedings{talmor2018web,
  title={The web as a knowledge-base for answering complex questions},
  author={Talmor, Alon and Berant, Jonathan},
  booktitle={ACL},
  year={2018}
}

@inproceedings{chang2022webqa,
  title={Webqa: Multihop and multimodal qa},
  author={Chang, Yingshan and Narang, Mridu and Suzuki, Hisami and Cao, Guihong and Gao, Jianfeng and Bisk, Yonatan},
  booktitle={CVPR},
  year={2022}
}

@inproceedings{ma2024mmlongbench,
  title={Mmlongbench-doc: Benchmarking long-context document understanding with visualizations},
  author={Ma, Yubo and Zang, Yuhang and Chen, Liangyu and Chen, Meiqi and Jiao, Yizhu and Li, Xinze and Lu, Xinyuan and Liu, Ziyu and Ma, Yan and Dong, Xiaoyi and others},
  booktitle={NeurIPS},
  year={2024}
}

@article{modarressi2025nolima,
  title={Nolima: Long-context evaluation beyond literal matching},
  author={Modarressi, Ali and Deilamsalehy, Hanieh and Dernoncourt, Franck and Bui, Trung and Rossi, Ryan A and Yoon, Seunghyun and Sch{\"u}tze, Hinrich},
  journal={arXiv preprint arXiv:2502.05167},
  year={2025}
}

@inproceedings{bai2024longbench,
  title={Longbench: A bilingual, multitask benchmark for long context understanding},
  author={Bai, Yushi and Lv, Xin and Zhang, Jiajie and Lyu, Hongchang and Tang, Jiankai and Huang, Zhidian and Du, Zhengxiao and Liu, Xiao and Zeng, Aohan and Hou, Lei and others},
  booktitle={ACL},
  year={2024}
}

@article{zhou2024causalbench,
  title={Causalbench: A comprehensive benchmark for causal learning capability of llms},
  author={Zhou, Yu and Wu, Xingyu and Huang, Beicheng and Wu, Jibin and Feng, Liang and Tan, Kay Chen},
  journal={arXiv preprint arXiv:2404.06349},
  year={2024}
}

@article{wu2024cofca,
  title={Cofca: A step-wise counterfactual multi-hop qa benchmark},
  author={Wu, Jian and Yang, Linyi and Wang, Zhen and Okumura, Manabu and Zhang, Yue},
  journal={arXiv preprint arXiv:2402.11924},
  year={2024}
}

@inproceedings{yu2023ifqa,
  title={Ifqa: A dataset for open-domain question answering under counterfactual presuppositions},
  author={Yu, Wenhao and Jiang, Meng and Clark, Peter and Sabharwal, Ashish},
  booktitle={EMNLP},
  year={2023}
}

@article{liu2023agentbench,
  title={Agentbench: Evaluating llms as agents},
  author={Liu, Xiao and Yu, Hao and Zhang, Hanchen and Xu, Yifan and Lei, Xuanyu and Lai, Hanyu and Gu, Yu and Ding, Hangliang and Men, Kaiwen and Yang, Kejuan and others},
  journal={arXiv preprint arXiv:2308.03688},
  year={2023}
}

@inproceedings{padmakumar2022teach,
  title={Teach: Task-driven embodied agents that chat},
  author={Padmakumar, Aishwarya and Thomason, Jesse and Shrivastava, Ayush and Lange, Patrick and Narayan-Chen, Anjali and Gella, Spandana and Piramuthu, Robinson and Tur, Gokhan and Hakkani-Tur, Dilek},
  booktitle={AAAI},
  year={2022}
}

@inproceedings{valmeekam2023planbench,
  title={Planbench: An extensible benchmark for evaluating large language models on planning and reasoning about change},
  author={Valmeekam, Karthik and Marquez, Matthew and Olmo, Alberto and Sreedharan, Sarath and Kambhampati, Subbarao},
  booktitle={NeurIPS},
  year={2023}
}

@inproceedings{akbari2021vatt,
  title={Vatt: Transformers for multimodal self-supervised learning from raw video, audio and text},
  author={Akbari, Hassan and Yuan, Liangzhe and Qian, Rui and Chuang, Wei-Hong and Chang, Shih-Fu and Cui, Yin and Gong, Boqing},
  booktitle={NeurIPS},
  year={2021}
}

@inproceedings{wasim2023vita,
  title={Vita-clip: Video and text adaptive clip via multimodal prompting},
  author={Wasim, Syed Talal and Naseer, Muzammal and Khan, Salman and Khan, Fahad Shahbaz and Shah, Mubarak},
  booktitle={CVPR},
  year={2023}
}

@inproceedings{chen2026learning,
  title={Learning compact video representations for efficient long-form video understanding in large multimodal models},
  author={Chen, Yuxiao and Wang, Jue and Zhang, Zhikang and Yi, Jingru and Zhang, Xu and Zou, Yang and Cai, Zhaowei and Yuan, Jianbo and Li, Xinyu and Yang, Hao and others},
  booktitle={WACV},
  year={2026}
}

@article{wu2024density,
  title={Density-based user representation using Gaussian process regression for multi-interest personalized retrieval},
  author={Wu, Haolun and Meshi, Ofer and Zoghi, Masrour and Diaz, Fernando and Liu, Xue and Boutilier, Craig and Karimzadehgan, Maryam},
  journal={Advances in Neural Information Processing Systems},
  volume={37},
  pages={52568--52594},
  year={2024}
}

@inproceedings{sadat2024co,
  title={Co-training for Low Resource Scientific Natural Language Inference},
  author={Sadat, Mobashir and Caragea, Cornelia},
  booktitle={Proceedings of the 62nd Annual Meeting of the Association for Computational Linguistics (Volume 1: Long Papers)},
  pages={2538--2550},
  year={2024}
}

@inproceedings{zhao2022towards,
  title={Towards efficient dialogue pre-training with transferable and interpretable latent structure},
  author={Zhao, Xueliang and Liu, Lemao and Fu, Tingchen and Shi, Shuming and Zhao, Dongyan and Yan, Rui},
  booktitle={Proceedings of the 2022 Conference on Empirical Methods in Natural Language Processing},
  pages={10051--10063},
  year={2022}
}

@inproceedings{li2023behavior1k,
  title={BEHAVIOR-1K: A Benchmark for Embodied AI with 1,000 Everyday Activities and Realistic Simulation},
  author={Li, Chengshu and others},
  booktitle={Proceedings of The 7th Conference on Robot Learning (CoRL)},
  year={2023}
}

@inproceedings{savva2019habitat,
  title={Habitat: A Platform for Embodied AI Research},
  author={Savva, Manolis and Kadian, Abhishek and Maksymets, Oleksandr and Zhao, Yili and Wijmans, Erik and Jain, Bhavana and Straub, Julian and Liu, Jia and Koltun, Vladlen and Malik, Jitendra and others},
  booktitle={Proceedings of the IEEE/CVF International Conference on Computer Vision (ICCV)},
  pages={9339--9347},
  year={2019}
}
\clearpage
\begin{center}
    {\huge \textbf{Supplementary Material}}
    \vspace{1cm}
\end{center}

\setcounter{section}{0}
\renewcommand{\thesection}{\Alph{section}}
\section{Details of Dataset Construction}

\subsection{Procedural Synthesis of Environments}
Tables~\ref{tab:scenario1}--\ref{tab:scenario10} summarize the scenarios used in our benchmark. Each scenario specifies a structured environment configuration, including the environment typology, the set of rooms, and the objects in each room. In addition, the scenario defines a predefined trajectory that the agent follows across rooms, as well as spatial relations such as adjacency and opposite room pairs, and vertical room pairs in multi-story environments.

Because the dataset should cover a broad range of situations rather than a small set of repetitive layouts, we made significant efforts to increase scenario diversity during construction. In particular, we varied room compositions, object distributions, traversal orders, and inter-room relations so that different scenarios would induce different spatial structures and reasoning requirements. The ten scenarios in our benchmark were designed based on this principle, with each one capturing a distinct combination of layout, object arrangement, and movement pattern. By explicitly specifying these components, each scenario provides a well-defined spatial layout and interaction structure. Across the benchmark, the scenarios vary in room compositions, object distributions, trajectories, and room relationships, allowing the dataset to cover diverse spatial configurations and reasoning situations. All scenarios were manually designed to ensure careful control over environment layouts and task conditions. This manual construction process allowed us to deliberately design environments that contain meaningful spatial relationships and object configurations, enabling targeted evaluation of spatial and temporal reasoning abilities.

\subsection{Human-in-the-Loop Annotation}
\subsubsection{Human-based Query Construction.}
The queries in our benchmark were constructed through a human-in-the-loop annotation process. In total, ten annotators participated in the query design stage. Each pair of annotators was assigned two scenarios, resulting in five annotation teams that collectively covered all ten scenarios. For each scenario, annotators created 30 queries, yielding a total of 300 queries across the dataset. All annotators were qualified graduate students with demonstrated research experience in computer vision. Specifically, each annotator had published at least one paper in a top-tier computer vision conference, ensuring familiarity with visual reasoning tasks and dataset construction practices.

The annotation process was organized to encourage scenario understanding and consistency. Each pair of annotators first designed the scenario for their assigned environments, specifying the room composition, agent trajectory, and the spatial relations. After establishing the scenario, they reviewed the corresponding environment layout and video trajectory to ensure consistency between the designed configuration and the rendered video. Based on this process, the annotators collaboratively designed 30 queries per scenario that target different aspects of spatial and temporal reasoning, including object relationships, navigation paths, and room-level spatial structures. Constructing such queries requires annotators to understand both the spatial configuration of the environment and the temporal sequence of the agent's movement, which makes the task substantially more demanding than writing simple object-centric questions. Throughout this process, the annotators were instructed to avoid trivial questions and instead focus on queries that require meaningful reasoning over the observed environment and temporal context.

\subsubsection{Human-based Query Validation.}
After the initial query construction stage, we conducted an additional human-based validation process to ensure that all queries were consistent with the video trajectories and could be correctly answered from the provided observations. The validation was performed by qualified annotators under the same criteria as those used for query construction. To avoid self-validation, annotators who constructed queries for a given scenario were assigned to validate different scenarios.

During validation, the annotators attempted to answer each query by carefully reviewing the rendered egocentric video and the underlying scenario configuration. As part of this process, they also cross-checked the floor plans they had independently inferred from the video to verify whether their spatial understanding was consistent. If a query was found to be ambiguous, inconsistent with the environment layout, or not fully resolvable from the video evidence, the annotators revisited the query and refined it through discussion. This iterative process often required multiple rounds of verification and revision, as designing questions that are both non-trivial and strictly answerable from a single video trajectory is inherently challenging.

\subsubsection{Compensation.}

Due to the iterative inspection and discussion required during query construction and validation, the process was time-intensive. Each scenario was handled by a team of two annotators who collaboratively designed and validated the queries. Completing a single scenario typically requires approximately two to three weeks. To compensate for this substantial effort, each annotator received approximately USD 350 per scenario.

\section{Details of Evaluation}

\subsection{Model List}
Table~\ref{tab:models} reports the SOTA MLLM versions used in our experiments. GPT, Gemini, and Claude were accessed through their official APIs, whereas the remaining models were accessed either through the OpenRouter api call or from publicly available Hugging Face repositories.
\setcounter{table}{4}
\begin{table}[t]
    \centering
    \scriptsize
    \begin{tabularx}{0.5\textwidth}{ >{\raggedright\arraybackslash}X >{\hsize=1.5\hsize\raggedright\arraybackslash}X }
        \toprule
        \textbf{Model Family} & \textbf{Model Name} \\
        \midrule
        GPT     & \texttt{gpt-5.2} \\
        Gemini  & \texttt{gemini-3-flash-preview} \\
                & \texttt{gemini-3-pro-preview} \\
        Claude  & \texttt{claude-haiku-4-5-20251001-v1:0} \\
                & \texttt{claude-sonnet-4-5-20250929-v1:0} \\
        \midrule
        Qwen    & \texttt{qwen3-vl-2b-instruct} \\
                & \texttt{qwen3-vl-8b-instruct} \\
                & \texttt{qwen3-vl-32b-instruct} \\
                & \texttt{qwen3-vl-235b-a22b-instruct} \\
        Intern  & \texttt{InternVL3\_5-2B-Instruct} \\
                & \texttt{InternVL3\_5-8B-Instruct} \\
        Mistral & \texttt{mistral-small-3.2-24b-instruct} \\ 
        Google & \texttt{gemma-3-27b-it} \\ 
        z-ai & \texttt{glm-4.6v} \\ 
        \bottomrule
    \end{tabularx}
    \caption{Set of proprietary and open-source multimodal large language models (MLLMs) evaluated on our benchmark.}
    \label{tab:models}
\end{table}

\subsection{Computational Resources}
For proprietary models, we accessed \texttt{GPT-5.2} via OpenAI, \texttt{Gemini-3-Flash} and \texttt{Gemini-3-Pro} via Gemini, and \texttt{Claude 4.5 Haiku} and \texttt{Claude 4.5 Sonnet} via Anthropic. Among the open-source models, \texttt{Qwen3-VL-2B}, \texttt{InternVL3.5-2B}, and \texttt{InternVL3.5-8B} were obtained from Hugging Face and served on GPU servers using 4 to 8 NVIDIA L40S GPUs, depending on the model configuration. The remaining models were accessed through the OpenRouter api.

\subsection{Additional Prompts}
We provide the full prompts used in our experiments for transparency and reproducibility. Table~\ref{tab:question_template} presents the detailed questions and answer choices for each abstractive spatiotemporal reasoning task, while the extractive tasks templates follow those introduced in VSI-Bench~\citep{yang2025thinking}. In all cases, the question text is combined with the corresponding system instruction and provided to the MLLM together with the input video.

Table~\ref{tab:mcq_na} reports the system instruction used for the MCQ and NA settings. This instruction requires the model to return only the selected choice for MCQ tasks or the final numeric answer for NA tasks, so that performance can be evaluated using exact-match accuracy.

Table~\ref{tab:llm_judge} presents the prompt used to evaluate outputs in the free-form generation setting. We conduct free-form evaluation on three tasks: Memory--Action, Map Direction, and Map Scale. Among these, Map Direction and Map Scale produce closed-ended answers that can be evaluated using exact-match accuracy. In contrast, Memory-Action requires open-ended responses. For this task, we evaluate the correctness of model outputs using an LLM-as-judge with the prompt shown in Table~\ref{tab:llm_judge}, where \texttt{gemini-3-flash-preview} is used as the judge.

Table~\ref{tab:floor_plan} presents the prompt used for the analysis of floor-plan prediction. Based on the prompt and the input video, the MLLM is asked to infer the spatial layout of the environment and generate a floor-plan description. The outputs are then transformed into a visual representation by manually mapping the predicted layout onto a 10 $\times$ 10 grid. In this process, the authors directly color-code each grid cell to reflect the inferred room structure, allowing visual comparison between the predicted floor plans and the ground-truth layouts.

\subsection{Evaluation Statistics}
To assess the stability of the reported results, we additionally conduct statistical analyses over five independent runs. We report the mean performance in the \textbf{Table 3} in the main paper, while the corresponding standard deviations, 95\% confidence intervals, and p-values are provided in Table~\ref{tab:abs_stats} and~\ref{tab:ext_stats}.

\smallskip
 \noindent\textbf{Standard Deviation.} While most tasks exhibit relatively small standard deviations, some tasks show larger variance across runs. This is mainly due to the limited number of evaluation instances and the stochastic nature of LLM decoding, which can introduce variability in multi-step reasoning outcomes. Nevertheless, the variation in overall average performance remains modest, and the relative ranking of models is largely consistent across runs.

\smallskip\smallskip
\noindent\textbf{Confidence Interval.} The 95\% confidence intervals are generally narrow across models, indicating that the reported results are stable over repeated runs. Wider intervals observed in a few task-specific cases are consistent with the larger standard deviations in those settings.

\smallskip\smallskip
\noindent\textbf{Significance Test.} Statistical significance is evaluated using paired two-sided t-tests across runs. Following prior work~\citep{zhao2022towards, sadat2024co, wu2024density}, we adopt the strongest-performing model as the reference baseline, which provides a strict and interpretable point of comparison. Under this criterion, \texttt{Gemini-3-Flash} serves as the baseline for the abstractive evaluation, while \texttt{Gemini-Pro} serves as the baseline for the extractive evaluation. We compute p-values only for the average score, as our primary goal is overall benchmark-level model comparison. 

\section{Discussion}

\subsection{Synthetic-to-Real Domain Gap}
A notable characteristic of our benchmark is that it is built in a synthetic environment. We emphasize this choice as a deliberate design decision for constructing realistic yet controllable evaluation scenarios. In our pipeline, the environments are not produced by generative synthesis, which could introduce additional artifacts or hallucinated structures. Instead, the full layouts are manually designed by humans in \texttt{SketchUp} and then rendered into egocentric videos with \texttt{Enscape}.

This design choice is also consistent with a broader line of work that adopts simulated environments when precise control over scene structure, actions, and evaluation conditions is required. Prior benchmarks, such as VisEscape~\citep{lim2025visescape} construct virtual environments tailored to target scenarios for dataset creation and evaluation. More broadly, simulation platforms and procedurally generated environments have been widely used in embodied AI to support diverse yet controllable evaluation settings~\citep{savva2019habitat, li2023behavior1k}. In this sense, although our benchmark is not collected from the physical world, it is grounded in realistic human-designed layouts and rendered with tools already used in professional practice, while still retaining the controllability that is difficult to achieve with real-world recordings.

We further validate the realism and suitability of this setup through human evaluation. Human participants were included as a baseline and achieved high accuracy on the benchmark, showing that the tasks are understandable and solvable for people. This suggests that our benchmark does not merely reflect artifacts arising from the synthetic environment, but instead captures reasoning problems that remain meaningful under human judgment. Together, these observations support the use of our benchmark as a valid testbed for the controlled evaluation of the targeted reasoning abilities.

\subsection{Scalability and Automation of the Pipeline}

Although our dataset is not large in scale compared to typical datasets, this is primarily due to the substantial cost and effort required to construct each scenario in our pipeline. Our environments are manually designed using professional 3D modeling and rendering tools. In particular, the environments are created in \texttt{SketchUp} and rendered into egocentric videos using \texttt{Enscape}. These tools incur licensing costs of approximately USD 600 each, and the environment construction process itself requires human labor for layout design, object placement, trajectory planning, and video rendering. In our current setup, human annotators were compensated for this work at approximately USD 350 per scenario (totally USD 3,500), and the full design and production process takes roughly two weeks for a single scenario.

Despite this relatively high construction cost, we argue that the dataset remains meaningful for evaluating spatial reasoning in multimodal models. First, each scenario contains a rich spatial configuration involving multiple rooms, objects, and trajectories, resulting in a complex reasoning environment rather than isolated static scenes. Second, when compared with related benchmarks that construct controlled environments, the scale of our dataset is comparable or larger. For example, VisEscape~\citep{lim2025visescape} constructs its dataset from 20 distinct rooms treated as separate environments, whereas our benchmark consists of 10 houses with an average of 6.7 rooms per scenario (totally 67 rooms). As a result, the total number of rooms and spatial configurations in our benchmark exceeds those of comparable scenario-based datasets.

These observations suggest that, while the dataset size is limited by the cost of high-quality environment construction, the resulting benchmark still provides a sufficiently rich and diverse set of scenarios for evaluating spatial reasoning capabilities in multimodal models.
\clearpage

\begin{table}[p]
    \centering
    
    \scriptsize 
    \begin{tabularx}{\linewidth}{p{2.8cm} X}
        \toprule
        \textbf{Task} & \textbf{Question \& Choices} \\
        \midrule
        \textbf{Memory--Action} & Based on the objects visible in the \textcolor{blue}{\{sequence\}} room you entered, what activity is possible in that room? \\
        & \textbf{Choices:} (A) Cooking (B) Sleeping (C) Working (D) Bathing (E) Dining \\
        \addlinespace[1ex]
        \textbf{Map Direction} & Given that \textcolor{blue}{\{kitchen\}} is east of the \textcolor{blue}{\{bedroom\}}, what is the direction of the \textcolor{blue}{\{study\}} relative to the \textcolor{blue}{\{bathroom\}}? \\
        & \textbf{Choices:} (A) East (B) West (C) South (D) North \\
        \addlinespace[1ex]
        \textbf{Map Scale} & Given that the Squared Distance between the \textcolor{blue}{\{bedroom\}} and \textcolor{blue}{\{living room\}} is 1. Based on this scale, what is the relative Squared Distance between the \textcolor{blue}{\{bathroom\}} and the \textcolor{blue}{\{kitchen\}}? \\
        & \textbf{Choices:} (A) 1 (B) 2 (C) 4 (D) 5 (E) 8 \\
        \addlinespace[1ex]
        \textbf{Simulation} & Based on the egocentric video, which of the given top-down layouts correctly represents the scene? \\
        & \textbf{Choices:} \\
        & (A) The \textcolor{blue}{\{bathroom\}} is adjacent to the \textcolor{blue}{\{living room\}}. \\
        & (B) To get from the \textcolor{blue}{\{bedroom\}} to the \textcolor{blue}{\{study\}}, you have to pass through \hspace*{1.2em} \textcolor{blue}{\{2 plants\}}.\\
        & (C) The \textcolor{blue}{\{dining room\}} is located in the corner. \\
        & (D) The \textcolor{blue}{\{kitchen\}} is located between the \textcolor{blue}{\{bedroom\}} and \textcolor{blue}{\{study\}}.\\
        & (E) There is only one way from the \textcolor{blue}{\{bathroom\}} to the \textcolor{blue}{\{living room\}}.\\
        \addlinespace[1ex]
        \textbf{Global Counting} & How many \textcolor{blue}{\{cups\}} are observed across all rooms in the video? \\
        \bottomrule
    \end{tabularx}
    \caption{Question Templates for Abstractive Spatiotemporal Reasoning. Note that the five answer choices shown here are illustrative examples. \textcolor{blue}{Blue} text indicates placeholders for rooms, objects, and counts/order, which may vary depending on the scenario.}
    \label{tab:question_template}

    \vspace{2em}

    \begin{tcolorbox}[
        colback=gray!5,
        colframe=black!75,
        title=MCQ \& NA,
        fonttitle=\bfseries\small,
        left=10pt, right=10pt, top=8pt, bottom=8pt,
        arc=2pt,
        width=0.95\linewidth
    ]
        \scriptsize
        \textbf{Question:} \\ 
        \texttt{Question\_Template[Task\_Name]} 
        
        \vspace{0.8em}
        \textbf{System Instruction:} \\
        Provide only your answer without any explanation. If it's a multiple-choice question, respond with only the letter (A, B, C, D, or E). If it's a counting question, respond with only the number.

        \vspace{0.8em}
        \textbf{Final Prompt:} \\
        \texttt{Question + System Instruction}
    \end{tcolorbox}
    \caption{Prompt design for MCQ/NA section.}
    \label{tab:mcq_na}
\end{table}
\clearpage
\begin{table}[p]
    \centering
    \scriptsize 
    \begin{tcolorbox}[
        colback=gray!5,
        colframe=black!75,
        title=LLM-Judge for Memory--Action,
        fonttitle=\bfseries\small,
        left=10pt, right=10pt, top=8pt, bottom=8pt,
        arc=2pt,
        width=0.95\linewidth
    ]
    
    \textbf{Input:} \\
    Target room: \textcolor{blue}{\{target room\}}\\
    Model\_answer: \textcolor{blue}{\{model\_answer\}}
    
    \vspace{1em}
    
    \textbf{System Instruction:} \\
    The question asks what activity is possible using objects visible in the target room. The model's answer may list multiple activities. Respond only ``correct'' if all activities in the model's answer can be done in the target room. Otherwise, respond only ``incorrect''.
    
    \end{tcolorbox}
    \caption{Prompt design for LLM-Judge section.}
    \label{tab:llm_judge}

    \vspace{2em}

    \begin{tcolorbox}[
        colback=gray!5,
        colframe=black!75,
        title=Floor-Plan Prediction,
        fonttitle=\bfseries\small,
        left=10pt, right=10pt, top=8pt, bottom=8pt,
        arc=2pt,
        width=0.95\linewidth
    ]
    
    \textbf{System Instruction:} \\
    Watch the video and reconstruct the floor plan on a $10 \times 10$ grid. Output strictly in JSON format.
    
    \vspace{1em}

    \textbf{Labels:}\\
    All rooms (including hallway and ``none'')

    \vspace{1em}
    
    \textbf{Coordinate System:} \\
    Assume the camera's start position in frame 1 is exactly at the start point. \\
    Start: (0, 0) | Bounds: (-5, 0), (5, 0), (-5, 10), (5, 10)
    
    \vspace{1em}

    \textbf{Output Format (Example):} \\
    \begin{minipage}{\linewidth}
    \scriptsize
    \begin{verbatim}
    {
    "reasoning": "Briefly trace the trajectory.",
    "grid": [
        ["none", "kitchen", "kitchen", ...], // y=9
        ["none", "kitchen", "kitchen", ...], // y=8
        ...
        ["none", "living room", "none", ...]] // y=0
    }
    \end{verbatim}
    \end{minipage}
    \end{tcolorbox}
    \caption{Prompt design for floor-plan prediction section.}
    \label{tab:floor_plan}
\end{table}
\clearpage

\begin{table}[p]
    \centering
    \scriptsize
    \begin{tabularx}{\textwidth}{X l cccccc}
        \toprule
        \textbf{Model} & \textbf{Stat} & \textbf{Mem.} & \textbf{MDir.} & \textbf{MSca.} & \textbf{Sim.} & \textbf{Glob.} & \textbf{Avg.} \\
        \midrule
        \multirow{3}{*}{GPT-5.2} 
        & mean $\pm$ \tiny{std}   & 38.0 $\pm$ \tiny{3.0} & 26.0 $\pm$ \tiny{6.0} & 34.0 $\pm$ \tiny{7.2} & 29.3 $\pm$ \tiny{6.4} & 23.3 $\pm$ \tiny{6.7} & 30.1 $\pm$ \tiny{2.5} \\
        & 95\% CI & [34.3,41.7] & [18.6,33.4] & [25.0,43.0] & [21.4,37.3] & [15.1,31.6] & [27.0,33.3] \\
        & p-value & - & - & - & - & - & 0.0085 \\
        \midrule
        \multirow{3}{*}{Gemini-3 Flash} 
        & mean $\pm$ \tiny{std} & 60.7 $\pm$ \tiny{2.8} & 34.0 $\pm$ \tiny{4.9} & 24.0 $\pm$ \tiny{3.7} & 51.3 $\pm$ \tiny{6.9} & 31.3 $\pm$ \tiny{7.3} & 40.3 $\pm$ \tiny{2.8} \\
        & 95\% CI & [57.2,64.1] & [27.9,40.1] & [19.5,28.5] & [42.8,59.9] & [22.3,40.4] & [36.8,43.7] \\
        & p-value & - & - & - & - & - & - \\
        \midrule
        \multirow{3}{*}{Gemini-3 Pro} 
        & mean $\pm$ \tiny{std} & 52.0 $\pm$ \tiny{3.0} & 22.7 $\pm$ \tiny{7.6} & 22.0 $\pm$ \tiny{5.6} & 31.3 $\pm$ \tiny{6.5} & 20.7 $\pm$ \tiny{3.7} & 29.7 $\pm$ \tiny{2.5} \\
        & 95\% CI & [48.3,55.7] & [13.2,32.1] & [15.1,28.9] & [23.3,39.4] & [16.1,25.2] & [26.7,32.8] \\
        & p-value & - & - & - & - & - & 0.0050 \\
        \midrule
        \multirow{3}{*}{Claude Haiku} 
        & mean $\pm$ \tiny{std} & 19.3 $\pm$ \tiny{1.5} & 21.3 $\pm$ \tiny{4.5} & 11.3 $\pm$ \tiny{3.0} & 22.0 $\pm$ \tiny{1.8} & 2.7 $\pm$ \tiny{2.8} & 15.3 $\pm$ \tiny{1.5} \\
        & 95\% CI & [17.5,21.2] & [15.8,26.9] & [7.6,15.0] & [19.7,24.3] & [-0.8,6.1] & [13.5,17.2] \\
        & p-value & - & - & - & - & - & 0.0001 \\
        \midrule
        \multirow{3}{*}{Claude Sonnet} 
        & mean $\pm$ \tiny{std} & 30.0 $\pm$ \tiny{2.4} & 19.3 $\pm$ \tiny{4.3} & 19.3 $\pm$ \tiny{2.8} & 40.7 $\pm$ \tiny{4.3} & 11.3 $\pm$ \tiny{1.8} & 24.1 $\pm$ \tiny{1.4} \\
        & 95\% CI & [27.1,32.9] & [13.9,24.7] & [15.9,22.8] & [35.3,46.1] & [9.1,13.6] & [22.3,25.9] \\
        & p-value & - & - & - & - & - & 0.0001 \\
        \midrule
        \multirow{3}{*}{Qwen3-VL-2B} 
        & mean $\pm$ \tiny{std} & 26.0 $\pm$ \tiny{2.8} & 18.7 $\pm$ \tiny{5.1} & 19.3 $\pm$ \tiny{3.7} & 26.0 $\pm$ \tiny{8.0} & 6.7 $\pm$ \tiny{6.7} & 19.3 $\pm$ \tiny{1.6} \\
        & 95\% CI & [22.5,29.5] & [12.4,24.9] & [14.8,23.9] & [16.1,35.9] & [-1.6,14.9] & [17.3,21.4] \\
        & p-value & - & - & - & - & - & 0.0002 \\
        \midrule
        \multirow{3}{*}{Qwen3-VL-8B} 
        & mean $\pm$ \tiny{std} & 34.0 $\pm$ \tiny{1.5} & 22.0 $\pm$ \tiny{1.8} & 15.3 $\pm$ \tiny{1.8} & 34.0 $\pm$ \tiny{2.8} & 17.3 $\pm$ \tiny{4.3} & 24.5 $\pm$ \tiny{1.4} \\
        & 95\% CI & [32.1,35.9] & [19.7,24.3] & [13.1,17.6] & [30.5,37.5] & [11.9,22.7] & [22.8,26.2] \\
        & p-value & - & - & - & - & - & 0.0001 \\
        \midrule
        \multirow{3}{*}{Qwen3-VL-32B} 
        & mean $\pm$ \tiny{std} & 40.0 $\pm$ \tiny{0.0} & 26.0 $\pm$ \tiny{2.8} & 23.3 $\pm$ \tiny{3.3} & 42.7 $\pm$ \tiny{2.8} & 17.3 $\pm$ \tiny{2.8} & 29.9 $\pm$ \tiny{1.7} \\
        & 95\% CI & [40.0,40.0] & [22.5,29.5] & [19.2,27.5] & [39.2,46.1] & [13.9,20.8] & [27.7,32.0] \\
        & p-value & - & - & - & - & - & 0.0033 \\
        \midrule
        \multirow{3}{*}{Qwen3-VL-235B} 
        & mean $\pm$ \tiny{std} & 43.3 $\pm$ \tiny{2.4} & 16.7 $\pm$ \tiny{4.1} & 13.3 $\pm$ \tiny{2.4} & 46.7 $\pm$ \tiny{4.1} & 13.3 $\pm$ \tiny{2.4} & 26.7 $\pm$ \tiny{2.0} \\
        & 95\% CI & [40.4,46.3] & [11.6,21.7] & [10.4,16.3] & [41.6,51.7] & [10.4,16.3] & [24.2,29.1] \\
        & p-value & - & - & - & - & - & 0.0004 \\
        \midrule
        \multirow{3}{*}{InternVL3.5-2B} 
        & mean $\pm$ \tiny{std} & 28.0 $\pm$ \tiny{3.8} & 22.7 $\pm$ \tiny{4.9} & 20.0 $\pm$ \tiny{4.7} & 21.3 $\pm$ \tiny{3.8} & 14.0 $\pm$ \tiny{3.7} & 21.2 $\pm$ \tiny{1.3} \\
        & 95\% CI & [23.3,32.7] & [16.5,28.8] & [14.1,25.9] & [16.6,26.1] & [9.5,18.5] & [19.6,22.8] \\
        & p-value & - & - & - & - & - & 0.0002 \\
        \midrule
        \multirow{3}{*}{InternVL3.5-8B} 
        & mean $\pm$ \tiny{std} & 36.7 $\pm$ \tiny{11.1} & 24.0 $\pm$ \tiny{3.7} & 21.3 $\pm$ \tiny{7.3} & 26.0 $\pm$ \tiny{4.3} & 16.0 $\pm$ \tiny{8.6} & 24.8 $\pm$ \tiny{2.7} \\
        & 95\% CI & [22.9,50.4] & [19.5,28.5] & [12.3,30.4] & [20.6,31.4] & [5.3,26.7] & [21.4,28.2] \\
        & p-value & - & - & - & - & - & 0.0003 \\
        \midrule
        \multirow{3}{*}{Mistral-3.2-24B} 
        & mean $\pm$ \tiny{std} & 28.7 $\pm$ \tiny{3.8} & 16.7 $\pm$ \tiny{6.2} & 10.0 $\pm$ \tiny{2.4} & 18.7 $\pm$ \tiny{5.1} & 14.0 $\pm$ \tiny{2.8} & 17.6 $\pm$ \tiny{2.9} \\
        & 95\% CI & [23.9,33.4] & [8.9,24.4] & [7.1,12.9] & [12.4,24.9] & [10.5,17.5] & [14.1,21.1] \\
        & p-value & - & - & - & - & - & 0.0000 \\
        \midrule
        \multirow{3}{*}{Gemma-3-27B} 
        & mean $\pm$ \tiny{std} & 32.0 $\pm$ \tiny{1.8} & 24.0 $\pm$ \tiny{2.8} & 23.3 $\pm$ \tiny{2.4} & 28.7 $\pm$ \tiny{1.8} & 10.7 $\pm$ \tiny{1.5} & 23.7 $\pm$ \tiny{0.8} \\
        & 95\% CI & [29.7,34.3] & [20.5,27.5] & [20.4,26.3] & [26.4,30.9] & [8.8,12.5] & [22.8,24.7] \\
        & p-value & - & - & - & - & - & 0.0003 \\
        \midrule
        \multirow{3}{*}{GLM-4.6V-106B} 
        & mean $\pm$ \tiny{std} & 35.3 $\pm$ \tiny{1.8} & 18.0 $\pm$ \tiny{6.1} & 20.7 $\pm$ \tiny{2.8} & 37.3 $\pm$ \tiny{4.9} & 6.0 $\pm$ \tiny{4.9} & 23.5 $\pm$ \tiny{2.8} \\
        & 95\% CI & [33.1,37.6] & [10.5,25.5] & [17.2,24.1] & [31.2,43.5] & [-0.1,12.1] & [20.0,26.9] \\
        & p-value & - & - & - & - & - & 0.0008 \\
        \bottomrule
    \end{tabularx}
    \caption{\textbf{Abstractive evaluation statistics.} Mean performance and standard deviation (mean $\pm$ std), 95\% confidence intervals (CI), and p-values across five runs for the abstractive evaluation tasks. Statistical significance is evaluated using paired two-sided t-tests.}
    \label{tab:abs_stats}
\end{table}
\begin{table}[p]
    \centering
    \scriptsize
    \begin{tabularx}{\textwidth}{X l cccccc}
        \toprule
        \textbf{Model} & \textbf{Stat} & \textbf{Appr.} & \textbf{Dir.} & \textbf{Dist.} & \textbf{Plan.} & \textbf{Obj.} & \textbf{Avg.} \\
        \midrule
        \multirow{3}{*}{GPT-5.2} 
        & mean $\pm$ \tiny{std}   & 73.3 $\pm$ \tiny{6.7} & 26.0 $\pm$ \tiny{4.3} & 56.0 $\pm$ \tiny{4.9} & 18.7 $\pm$ \tiny{7.3} & 48.7 $\pm$ \tiny{6.9} & 44.5 $\pm$ \tiny{3.1} \\
        & 95\% CI & [65.1,81.6] & [20.6,31.4] & [49.9,62.1] & [9.6,27.7] & [40.1,57.2] & [40.6,48.4] \\
        & p-value & - & - & - & - & - & 0.0176 \\
        \midrule
        \multirow{3}{*}{Gemini-3 Flash} 
        & mean $\pm$ \tiny{std} & 93.3 $\pm$ \tiny{0.0} & 40.0 $\pm$ \tiny{7.5} & 62.7 $\pm$ \tiny{1.5} & 16.0 $\pm$ \tiny{2.8} & 38.0 $\pm$ \tiny{3.8} & 50.0 $\pm$ \tiny{2.4} \\
        & 95\% CI & [93.3,93.3] & [30.7,49.3] & [60.8,64.5] & [12.5,19.5] & [33.3,42.7] & [47.1,52.9] \\
        & p-value & - & - & - & - & - & 0.0341 \\
        \midrule
        \multirow{3}{*}{Gemini-3 Pro} 
        & mean $\pm$ \tiny{std} & 96.0 $\pm$ \tiny{4.3} & 44.7 $\pm$ \tiny{11.7} & 66.7 $\pm$ \tiny{4.7} & 26.0 $\pm$ \tiny{7.2} & 44.7 $\pm$ \tiny{3.8} & 55.6 $\pm$ \tiny{3.4} \\
        & 95\% CI & [90.6,101.4] & [30.2,59.2] & [60.8,72.5] & [17.0,35.0] & [39.9,49.4] & [51.4,59.8] \\
        & p-value & - & - & - & - & - & - \\
        \midrule
        \multirow{3}{*}{Claude Haiku} 
        & mean $\pm$ \tiny{std} & 34.0 $\pm$ \tiny{2.8} & 30.0 $\pm$ \tiny{3.3} & 26.7 $\pm$ \tiny{3.3} & 30.0 $\pm$ \tiny{4.1} & 40.0 $\pm$ \tiny{2.4} & 32.1 $\pm$ \tiny{1.7} \\
        & 95\% CI & [30.5,37.5] & [25.9,34.1] & [22.5,30.8] & [24.9,35.1] & [37.1,42.9] & [30.0,34.3] \\
        & p-value & - & - & - & - & - & 0.0000 \\
        \midrule
        \multirow{3}{*}{Claude Sonnet} 
        & mean $\pm$ \tiny{std} & 72.7 $\pm$ \tiny{2.8} & 24.7 $\pm$ \tiny{3.0} & 28.0 $\pm$ \tiny{5.1} & 13.3 $\pm$ \tiny{0.0} & 30.7 $\pm$ \tiny{2.8} & 33.9 $\pm$ \tiny{1.3} \\
        & 95\% CI & [69.2,76.1] & [21.0,28.4] & [21.7,34.3] & [13.3,13.3] & [27.2,34.1] & [32.3,35.5] \\
        & p-value & - & - & - & - & - & 0.0002\\
        \midrule
        \multirow{3}{*}{Qwen3-VL-2B} 
        & mean $\pm$ \tiny{std} & 82.0 $\pm$ \tiny{3.8} & 29.3 $\pm$ \tiny{10.9} & 42.7 $\pm$ \tiny{5.5} & 22.7 $\pm$ \tiny{7.2} & 40.0 $\pm$ \tiny{2.4} & 43.3 $\pm$ \tiny{4.2} \\
        & 95\% CI & [77.3,86.7] & [15.8,42.9] & [35.9,49.5] & [13.7,31.6] & [37.1,42.9] & [38.1,48.6] \\
        & p-value & - & - & - & - & - & 0.0004 \\
        \midrule
        \multirow{3}{*}{Qwen3-VL-8B} 
        & mean $\pm$ \tiny{std} & 80.0 $\pm$ \tiny{0.0} & 33.3 $\pm$ \tiny{4.1} & 54.0 $\pm$ \tiny{1.5} & 25.3 $\pm$ \tiny{1.8} & 34.0 $\pm$ \tiny{4.3} & 45.3 $\pm$ \tiny{0.7} \\
        & 95\% CI & [80.0,80.0] & [28.3,38.4] & [52.1,55.9] & [23.1,27.6] & [28.6,39.4] & [44.5,46.2] \\
        & p-value & - & - & - & - & - &  0.0026 \\
        \midrule
        \multirow{3}{*}{Qwen3-VL-32B} 
        & mean $\pm$ \tiny{std} & 90.0 $\pm$ \tiny{0.0} & 26.0 $\pm$ \tiny{2.8} & 53.3 $\pm$ \tiny{0.0} & 6.7 $\pm$ \tiny{0.0} & 51.3 $\pm$ \tiny{4.5} & 45.5 $\pm$ \tiny{1.4} \\
        & 95\% CI & [90.0,90.0] & [22.5,29.5] & [53.3,53.3] & [6.7,6.7] & [45.8,56.9] & [43.7,47.3] \\
        & p-value & - & - & - & - & - & 0.0021 \\
        \midrule
        \multirow{3}{*}{Qwen3-VL-235B} 
        & mean $\pm$ \tiny{std} & 93.3 $\pm$ \tiny{0.0} & 38.0 $\pm$ \tiny{3.8} & 54.0 $\pm$ \tiny{2.8} & 20.0 $\pm$ \tiny{6.2} & 43.3 $\pm$ \tiny{2.4} & 49.7 $\pm$ \tiny{1.8} \\
        & 95\% CI & [93.3,93.3] & [33.3,42.7] & [50.5,57.5] & [12.3,27.7] & [40.4,46.3] & [47.5,52.0] \\
        & p-value & - & - & - & - & - & 0.0500 \\
        \midrule
        \multirow{3}{*}{InternVL3.5-2B} 
        & mean $\pm$ \tiny{std} & 64.0 $\pm$ \tiny{4.3} & 30.7 $\pm$ \tiny{8.0} & 36.7 $\pm$ \tiny{7.5} & 20.0 $\pm$ \tiny{7.1} & 30.7 $\pm$ \tiny{7.6} & 36.4 $\pm$ \tiny{1.9} \\
        & 95\% CI & [58.6,69.4] & [20.8,40.5] & [27.4,45.9] & [11.2,28.8] & [21.2,40.1] & [34.0,38.8] \\
        & p-value & - & - & - & - & - & 0.0008 \\
        \midrule
        \multirow{3}{*}{InternVL3.5-8B} 
        & mean $\pm$ \tiny{std} & 73.3 $\pm$ \tiny{4.7} & 28.7 $\pm$ \tiny{6.9} & 36.0 $\pm$ \tiny{9.5} & 20.7 $\pm$ \tiny{5.5} & 32.0 $\pm$ \tiny{5.1} & 38.1 $\pm$ \tiny{3.5} \\
        & 95\% CI & [67.5,79.2] & [20.1,37.2] & [24.1,47.9] & [13.9,27.5] & [25.7,38.3] & [33.8,42.5] \\
        & p-value & - & - & - & - & - & 0.0011 \\
        \midrule
        \multirow{3}{*}{Mistral-3.2-24B} 
        & mean $\pm$ \tiny{std} & 28.7 $\pm$ \tiny{7.3} & 28.0 $\pm$ \tiny{6.9} & 40.0 $\pm$ \tiny{5.3} & 17.3 $\pm$ \tiny{6.0} & 40.7 $\pm$ \tiny{6.4} & 30.9 $\pm$ \tiny{2.4} \\
        & 95\% CI & [19.6,37.7] & [19.4,36.6] & [33.5,46.5] & [9.9,24.7] & [32.7,48.6] & [27.9,34.0] \\
        & p-value & - & - & - & - & - & 0.0000 \\
        \midrule
        \multirow{3}{*}{Gemma-3-27B} 
        & mean $\pm$ \tiny{std} & 55.3 $\pm$ \tiny{3.0} & 33.3 $\pm$ \tiny{0.0} & 28.0 $\pm$ \tiny{1.8} & 28.7 $\pm$ \tiny{1.8} & 49.3 $\pm$ \tiny{2.8} & 38.9 $\pm$ \tiny{0.8} \\
        & 95\% CI & [51.6,59.0] & [33.3,33.3] & [25.7,30.3] & [26.4,30.9] & [45.9,52.8] & [38.0,39.9] \\
        & p-value & - & - & - & - & - & 0.0006 \\
        \midrule
        \multirow{3}{*}{GLM-4.6V-106B} 
        & mean $\pm$ \tiny{std} & 35.3 $\pm$ \tiny{1.8} & 18.0 $\pm$ \tiny{6.1} & 20.7 $\pm$ \tiny{2.8} & 37.3 $\pm$ \tiny{4.9} & 6.0 $\pm$ \tiny{4.9} & 23.5 $\pm$ \tiny{2.8} \\
        & 95\% CI & [33.1,37.6] & [10.5,25.5] & [17.2,24.1] & [31.2,43.5] & [-0.1,12.1] & [20.0,26.9] \\
        & p-value & - & - & - & - & - & 0.0020\\
        \bottomrule
    \end{tabularx}
    \caption{\textbf{Extractive evaluation statistics.} Mean performance and standard deviation (mean $\pm$ std), 95\% confidence intervals (CI), and p-values across five runs for the extractive evaluation tasks. Statistical significance is evaluated using paired two-sided t-tests.}
    \label{tab:ext_stats}
\end{table}
\setcounter{figure}{5}

\begin{figure}[p] 
    \centering

    \includegraphics[width=0.8\linewidth]{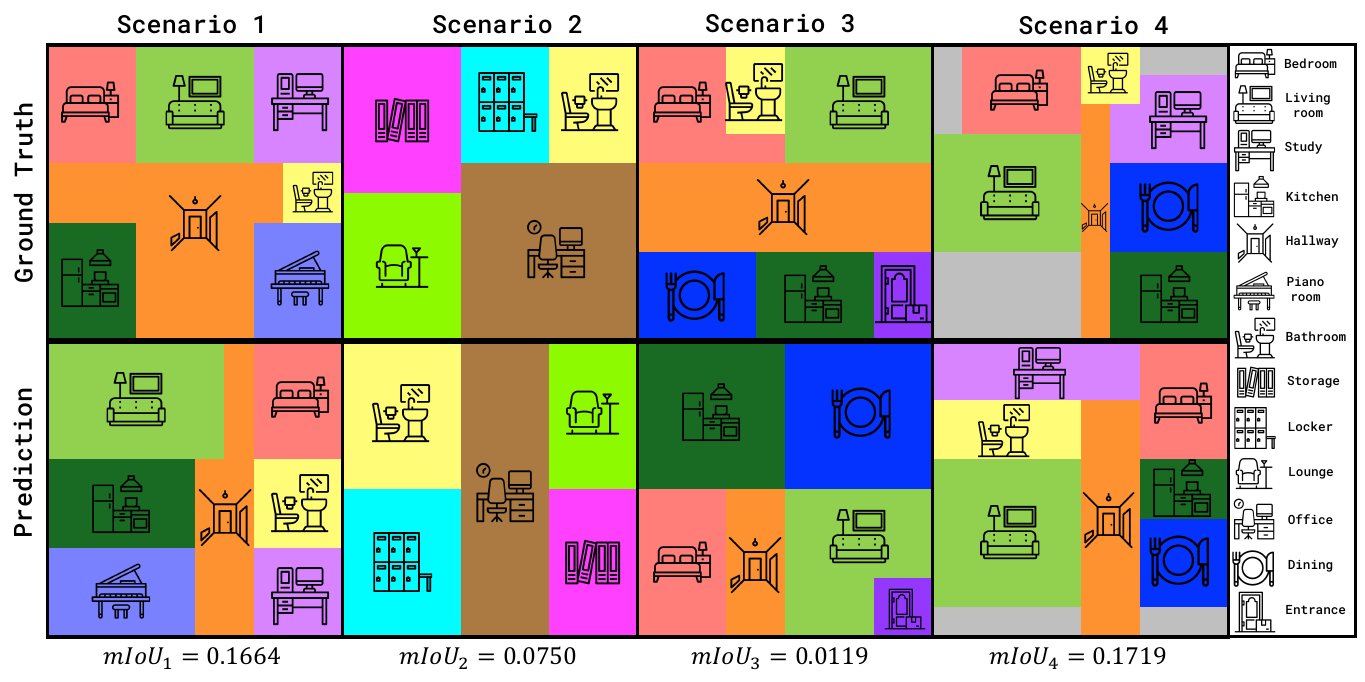}
    \caption{Gemini-3-pro floor-plan prediction}
    \label{fig:floor_plan1}
    
    \vspace{1cm}

    \includegraphics[width=0.8\linewidth]{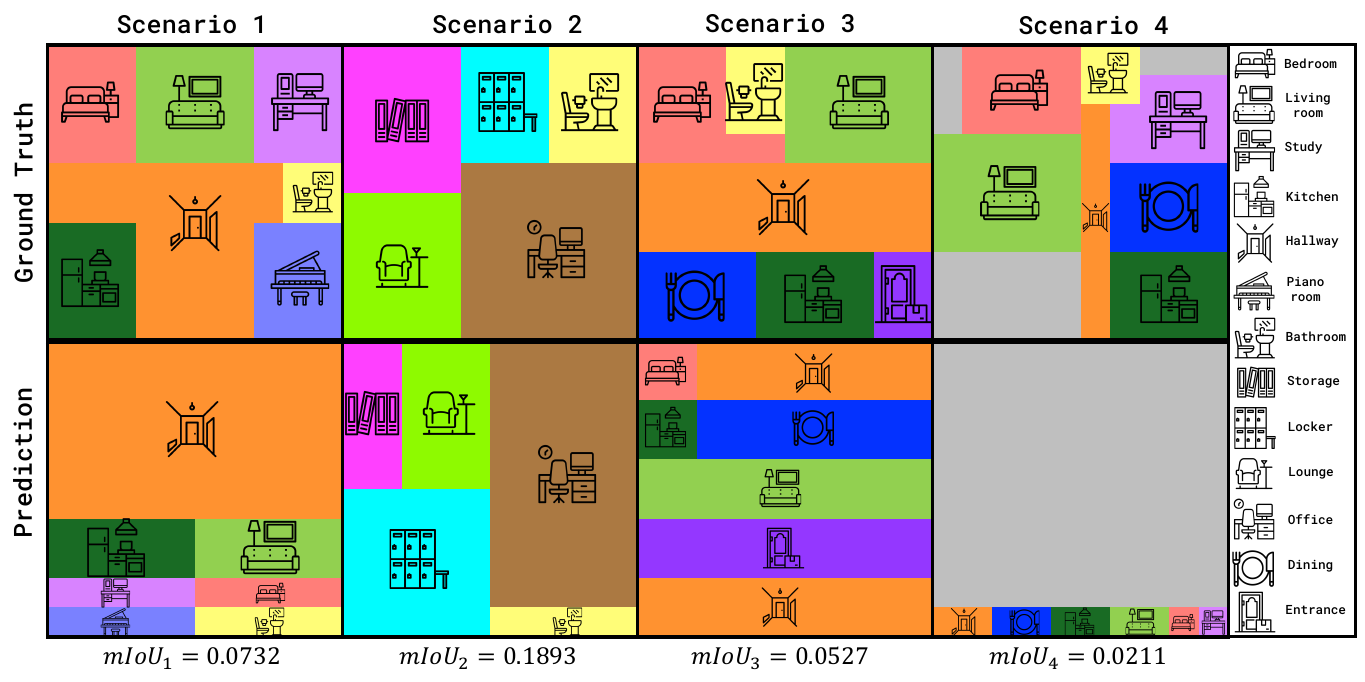}
    \caption{GPT-5.2 floor-plan prediction}
    \label{fig:floor_plan2}
    
    \vspace{1cm}

    \includegraphics[width=0.8\linewidth]{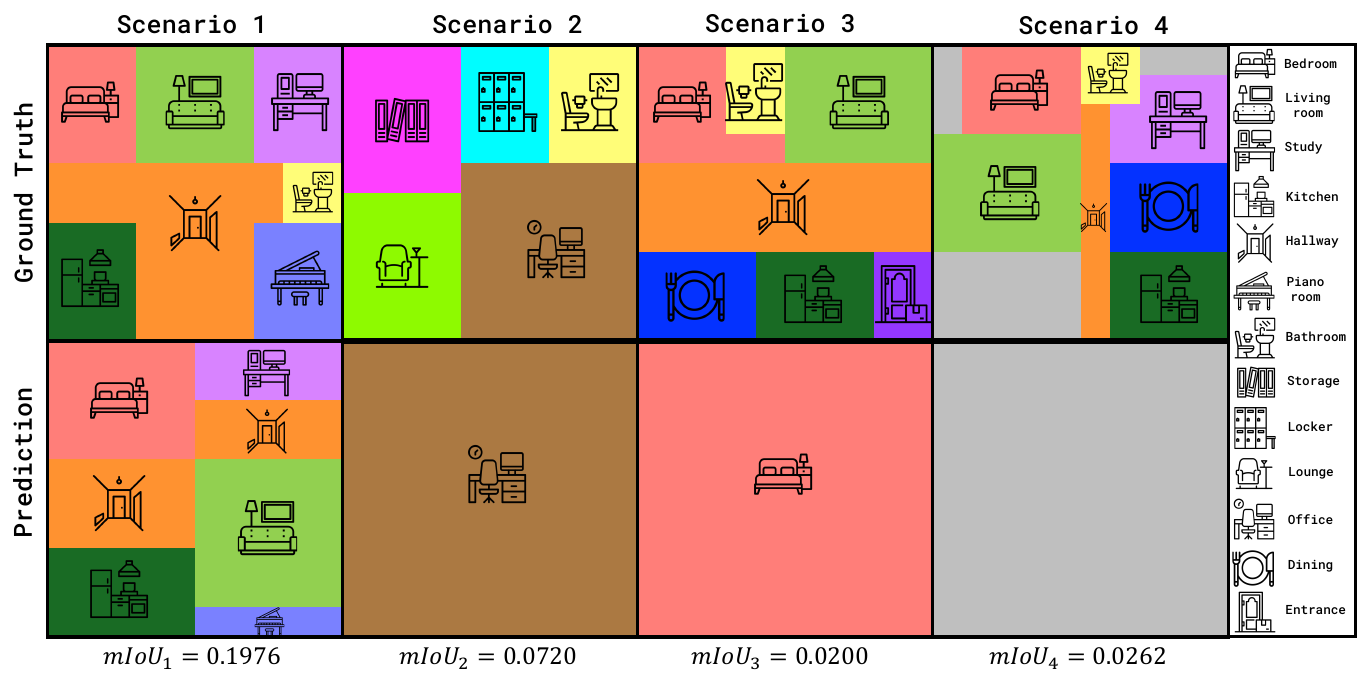}
    \caption{Qwen3-VL-235B-A22B floor-plan prediction}
    \label{fig:floor_plan3}
    
\end{figure}
\clearpage 
\begin{table}[p]
    \centering
    \scriptsize
    \begin{tabularx}{\linewidth}{p{3cm}X}
        \toprule
        \textbf{Category} & \textbf{Contents} \\
        \midrule
        Environment typology & Single-story residence \\
        \midrule
        \# of rooms & 6 \\
        \addlinespace[0.5ex]
        Types of rooms & bathroom, bedroom, kitchen, living room, piano room, study \\
        \midrule
        \multirow{6}{*}{Objects} 
        & \textbf{bathroom}: washstand, toilet \\ \addlinespace[0.5ex]
        & \textbf{bedroom}: wardrobe, lamp, cabinet, bed, books, pillows, windows \\ \addlinespace[0.5ex]
        & \textbf{kitchen}: fridge, cupboard, stove, sink, window, cups \\ \addlinespace[0.5ex]
        & \textbf{living room}: cabinet, books, window, TV, sofas, table, cups \\ \addlinespace[0.5ex]
        & \textbf{piano room}: piano, book, cup, window \\ \addlinespace[0.5ex]
        & \textbf{study}: bookshelf, computer, desk, chair, cup, windows \\ 
        \midrule
        Trajectory & kitchen $\rightarrow$ living room $\rightarrow$ bedroom $\rightarrow$ bathroom $\rightarrow$ study $\rightarrow$ piano room \\
        \midrule
        \multirow{2}{*}{Room Relations}
        & \textbf{Adjacency}: bedroom--living room, living room--study, study--bathroom, bathroom--piano room \\ \addlinespace[0.5ex]
        & \textbf{Opposite}: bedroom--kitchen, kitchen--piano room, piano room--study \\
        \bottomrule
    \end{tabularx}
    \caption{Scenario 1}
    \label{tab:scenario1}

    \vspace{2cm}

    \begin{tabularx}{\linewidth}{p{3cm}X}
        \toprule
        \textbf{Category} & \textbf{Contents} \\
        \midrule
        Environment typology & Single-story residence \\
        \midrule
        \# of rooms & 6 \\
        \addlinespace[0.5ex]
        Types of rooms & bathroom, bedroom 1, bedroom 2, dressroom, kitchen, living room \\
        \midrule
        \multirow{6}{*}{Objects} 
        & \textbf{bathroom}: washstand, toilet, window \\ \addlinespace[0.5ex]
        & \textbf{bedroom 1}: bed, pillows, telephone, handbag, wardrobe, window \\ \addlinespace[0.5ex]
        & \textbf{bedroom 2}: handbag, bed, pillow, plant, window \\ \addlinespace[0.5ex] 
        & \textbf{dressroom}: handbags, clothes, wardrobe, window \\ \addlinespace[0.5ex]
        & \textbf{kitchen}: microwave, pot, stove, oven, sink, vase, fruits, window, cupboard \\ \addlinespace[0.5ex]
        & \textbf{living room}: table, chairs, plants, sofa, TV, books, window \\
        \midrule
        Trajectory & kitchen $\rightarrow$ bathroom $\rightarrow$ bedroom 1 $\rightarrow$ living room $\rightarrow$ dressroom $\rightarrow$ bedroom 2 \\
        \midrule
        \multirow{2}{*}{Room Relations}
        & \textbf{Adjacency}: bathroom--bedroom 1, kitchen--bathroom, dressroom--bedroom 2 \\ \addlinespace[0.5ex]
        & \textbf{Opposite}: \{kitchen, bathroom, bedroom 1\}--\{dressroom, bedroom 2\} \\
        \bottomrule
    \end{tabularx}
    \caption{Scenario 2}
    \label{tab:scenario2}
\end{table}
\clearpage

\begin{table}[p]
    \centering
    \scriptsize
    \begin{tabularx}{\linewidth}{p{3cm}X}
        \toprule
        \textbf{Category} & \textbf{Contents} \\
        \midrule
        Environment typology & Single-story residence \\
        \midrule
        \# of rooms & 6 \\
        \addlinespace[0.5ex]
        Types of rooms & bathroom, bedroom, dining room, kitchen, living room, study \\
        \midrule
        \multirow{6}{*}{Objects} 
        & \textbf{bathroom}: washstand, bath, toilet, window \\ \addlinespace[0.5ex]
        & \textbf{bedroom}: bed, pillow, wardrobe, window \\ \addlinespace[0.5ex]
        & \textbf{dining room}: table, chairs, book, fireplace, window, door \\ \addlinespace[0.5ex]
        & \textbf{kitchen}: cupboards, windows, pot, stove, books, door \\ \addlinespace[0.5ex]
        & \textbf{living room}: TV, sofa, plants, wardrobe, book, windows, door \\ \addlinespace[0.5ex]
        & \textbf{study}: desk, chair, computer, lamp  \\
        \midrule
        Trajectory & dining room $\rightarrow$ kitchen $\rightarrow$ bedroom $\rightarrow$ study $\rightarrow$ living room $\rightarrow$ bathroom \\
        \midrule
        \multirow{2}{*}{Room Relations}
        & \textbf{Adjacency}: dining room--kitchen, living room--bathroom \\ \addlinespace[0.5ex]
        & \textbf{Opposite}: bedroom--study, kitchen--living room \\
        \bottomrule
    \end{tabularx}
    \caption{Scenario 3}
    \label{tab:scenario3}

    \vspace{2cm}

    \begin{tabularx}{\linewidth}{p{3cm}X}
        \toprule
        \textbf{Category} & \textbf{Contents} \\
        \midrule
        Environment typology & Single-story residence \\
        \midrule
        \# of rooms & 8 \\
        \addlinespace[0.5ex]
        Types of rooms & bathroom, bedroom 1, bedroom 2, dining room, homegym, kitchen, living room, study \\
        \midrule
        \multirow{8}{*}{Objects} 
        & \textbf{bathroom}: window, washstand, toilet \\ \addlinespace[0.5ex]
        & \textbf{bedroom 1}: bed, pillows, table, TV, carpet, plant, wardrobe, pictures, floor lamp \\ \addlinespace[0.5ex]
        & \textbf{bedroom 2}: bed, pillow, floor lamps, chairs, TV, picture, window \\ \addlinespace[0.5ex]
        & \textbf{dining room}: table, chairs, carpet, pictures, floor lamp, plants, windows \\ \addlinespace[0.5ex]
        & \textbf{homegym}: yoga mat, window, exercise machine, bench \\ \addlinespace[0.5ex]
        & \textbf{kitchen}: oven, plant, pot, stove \\ \addlinespace[0.5ex]
        & \textbf{living room}: sofas, TV, plants, table, carpet \\ \addlinespace[0.5ex]
        & \textbf{study}: desk, chair, computer, lamp, bookshelf, window \\
        \midrule
        Trajectory & bedroom 1 $\rightarrow$ bathroom $\rightarrow$ kitchen $\rightarrow$ study $\rightarrow$ dining room $\rightarrow$ homegym $\rightarrow$ bedroom 2 $\rightarrow$ living room \\
        \midrule
        \multirow{2}{*}{Room Relations}
        & \textbf{Adjacency}: homegym--bedroom 2--bathroom--bedroom 1, bedroom 1--kitchen--study, living room--dining room--study, homegym--living room  \\ \addlinespace[0.5ex]
        & \textbf{Opposite}: kitchen--living room, study--living room \\
        \bottomrule
    \end{tabularx}
    \caption{Scenario 4}
    \label{tab:scenario4}
\end{table}
\clearpage

\begin{table}[p]
    \centering
    \scriptsize
    \begin{tabularx}{\linewidth}{p{3cm}X}
        \toprule
        \textbf{Category} & \textbf{Contents} \\
        \midrule
        Environment typology & Office \\
        \midrule
        \# of rooms & 5 \\
        \addlinespace[0.5ex]
        Types of rooms & bathroom, file room, locker room, lounge, main office  \\
        \midrule
        \multirow{5}{*}{Objects} 
        & \textbf{bathroom}: washstand, toilet \\ \addlinespace[0.5ex]
        & \textbf{file room}: desk, chair, carpet, bookshelf, window, printer, prints, safe \\ \addlinespace[0.5ex]
        & \textbf{locker room}: lockers, carpet \\ \addlinespace[0.5ex]
        & \textbf{lounge}: sofas, table, tumbler, TV, microwave, fridge \\ \addlinespace[0.5ex]
        & \textbf{main office}: tumblers, computers, desks, chairs, carpets \\
        \midrule
        Trajectory & main office $\rightarrow$ bathroom $\rightarrow$ locker room $\rightarrow$ lounge $\rightarrow$ file room \\
        \midrule
        \multirow{2}{*}{Room Relations}
        & \textbf{Adjacency}: main office--\{locker room, bathroom, lounge\}, locker room--\{bathroom, file room\}, lounge--file room \\ \addlinespace[0.5ex]
        & \textbf{Opposite}: none \\
        \bottomrule
    \end{tabularx}
    \caption{Scenario 5}
    \label{tab:scenario5}

    \vspace{2cm}

    \begin{tabularx}{\linewidth}{p{3cm}X}
        \toprule
        \textbf{Category} & \textbf{Contents} \\
        \midrule
        Environment typology & Office \\
        \midrule
        \# of rooms & 6 \\
        \addlinespace[0.5ex]
        Types of rooms & bathroom, CEO room, conference room, locker room, lounge, main office \\
        \midrule
        \multirow{6}{*}{Objects} 
        & \textbf{bathroom}: washstand, toilet \\ \addlinespace[0.5ex]
        & \textbf{CEO room}: laptop, mouse, desk, chair, picture, printer, window \\ \addlinespace[0.5ex]
        & \textbf{conference room}: table, chairs, windows \\ \addlinespace[0.5ex]
        & \textbf{locker room}: lockers \\ \addlinespace[0.5ex]
        & \textbf{lounge}: sofas, coffee machine, vending machine, table \\ \addlinespace[0.5ex]
        & \textbf{main office}: desks, chairs, laptops, printer \\
        \midrule
        Trajectory & bathroom $\rightarrow$ lounge $\rightarrow$ conference room $\rightarrow$ CEO room $\rightarrow$ locker room $\rightarrow$ main office \\
        \midrule
        \multirow{2}{*}{Room Relations}
        & \textbf{Adjacency}: CEO room--conference room--lounge, lounge--bathroom, locker room--main office \\ \addlinespace[0.5ex]
        & \textbf{Opposite}: conference room--\{main office, locker room\} \\
        \bottomrule
    \end{tabularx}
    \caption{Scenario 6}
    \label{tab:scenario6}
\end{table}
\clearpage

\begin{table}[p]
    \centering
    \scriptsize
    \begin{tabularx}{\linewidth}{p{3cm}X}
        \toprule
        \textbf{Category} & \textbf{Contents} \\
        \midrule
        Environment typology & Single-story residence \\
        \midrule
        \# of rooms & 6 \\
        \addlinespace[0.5ex]
        Types of rooms & bathroom, bedroom, dining room, kitchen, living room, study \\
        \midrule
        \multirow{6}{*}{Objects} 
        & \textbf{bathroom}: window, towers, curtain, toilet \\ \addlinespace[0.5ex]
        & \textbf{bedroom}: bed, pillow, air conditioner, wardrobe, window \\ \addlinespace[0.5ex]
        & \textbf{dining room}: table, chairs, window \\ \addlinespace[0.5ex]
        & \textbf{kitchen}: microwave, stove, cupboards, fridge, window  \\ \addlinespace[0.5ex]
        & \textbf{living room}: sofa, table, air conditioner, window, TV \\ \addlinespace[0.5ex]
        & \textbf{study}: window, air conditioner, desk, chair, computer, bookshelf \\
        \midrule
        Trajectory & dining room $\rightarrow$ kitchen $\rightarrow$ living room $\rightarrow$ bedroom $\rightarrow$ bathroom $\rightarrow$ study \\
        \midrule
        \multirow{2}{*}{Room Relations}
        & \textbf{Adjacency}: living room--bedroom, bedroom--bathroom--study, study-dining room--kitchen \\ \addlinespace[0.5ex]
        & \textbf{Opposite}: dining room--living room, bedroom--study \\
        \bottomrule
    \end{tabularx}
    \caption{Scenario 7}
    \label{tab:scenario7}

    \vspace{2cm}

    \begin{tabularx}{\linewidth}{p{3cm}X}
        \toprule
        \textbf{Category} & \textbf{Contents} \\
        \midrule
        Environment typology & Multi-story residence \\
        \midrule
        \# of rooms & 6 \\
        \addlinespace[0.5ex]
        Types of rooms & bedroom 1, bedroom 2, kitchen \& dining room, homegym, living room, study \\
        \midrule
        \multirow{6}{*}{Objects} 
        & \textbf{bedroom 1}: desk, chair, window, bed, pillows \\ \addlinespace[0.5ex]
        & \textbf{bedroom 2}: bed, pillows, cabinet, window \\ \addlinespace[0.5ex]
        & \textbf{kitchen \& dining room}: fridge, sink, apples, eyeglasses, wines, stove, table, chairs, plates, windows \\ \addlinespace[0.5ex]
        & \textbf{homegym}: exercise machine, bench, cycle \\ \addlinespace[0.5ex]
        & \textbf{living room}: sofa, apples, plant, table, windows \\ \addlinespace[0.5ex]
        & \textbf{study}: desk, chair, computer, headphone, window, bookshelf \\
        \midrule
        Trajectory & kitchen \& dining room $\rightarrow$ bedroom 1 $\rightarrow$ study $\rightarrow$ bedroom 2 $\rightarrow$ living room $\rightarrow$ homegym \\
        \midrule
        \multirow{3}{*}{Room Relations}
        & \textbf{Adjacency}: living room--study--bedroom 1--kitchen \& dining room \\ \addlinespace[0.5ex]
        & \textbf{Opposite}:  study--bedroom 2 \\ \addlinespace[0.5ex]
        & \textbf{Vertical}: bedroom 2--homegym \\
        \bottomrule
    \end{tabularx}
    \caption{Scenario 8}
    \label{tab:scenario8}
\end{table}
\clearpage

\begin{table}[p]
    \centering
    \scriptsize
    \begin{tabularx}{\linewidth}{p{3cm}X}
        \toprule
        \textbf{Category} & \textbf{Contents} \\
        \midrule
        Environment typology & Multi-story residence \\
        \midrule
        \# of rooms & 8 \\
        \addlinespace[0.5ex]
        Types of rooms & bathroom 1, bathroom 2, bedroom 1, bedroom 2, dining room, dressroom, kitchen, living room\\
        \midrule
        \multirow{8}{*}{Objects} 
        & \textbf{bathroom 1}: washstand, cup, toilet, mirror, plant \\ \addlinespace[0.5ex]
        & \textbf{bathroom 2}: washstand, toilet, mirror \\ \addlinespace[0.5ex]
        & \textbf{bedroom 1}: bed, pillows, lamps, picture, vase, chair, window, curtain \\ \addlinespace[0.5ex]
        & \textbf{bedroom 2}: bed, pillow, window, cups, chair, bookshelf, curtain \\ \addlinespace[0.5ex]
        & \textbf{dining room}: table, chairs, cups, books \\ \addlinespace[0.5ex]
        & \textbf{dressroom}: wardrobe, window, vase \\ \addlinespace[0.5ex]
        & \textbf{kitchen}: fridge, washing machine, sink, stove, cupboard \\ \addlinespace[0.5ex]
        & \textbf{living room}: table, sofas, TV, vase, picture, lamp \\
        \midrule
        Trajectory & living room $\rightarrow$ dining room $\rightarrow$ kitchen $\rightarrow$ bedroom 1 $\rightarrow$ bathroom 1 $\rightarrow$ dressroom $\rightarrow$ bathroom 2 $\rightarrow$ bedroom 2\\
        \midrule
        \multirow{3}{*}{Room Relations}
        & \textbf{Adjacency}: bedroom--living room, living room--study, study--bathroom, bathroom--piano room \\ \addlinespace[0.5ex]
        & \textbf{Opposite}: bedroom--kitchen, kitchen--piano room, piano room--study \\ \addlinespace[0.5ex]
        & \textbf{Vertical}: living room--bedroom1--bedroom2, dining room--bathroom 1--bathroom 2\\
        \bottomrule
    \end{tabularx}
    \caption{Scenario 9}
    \label{tab:scenario9}

    \vspace{2cm}

    \begin{tabularx}{\linewidth}{p{3cm}X}
        \toprule
        \textbf{Category} & \textbf{Contents} \\
        \midrule
        Environment typology & Multi-story residence \\
        \midrule
        \# of rooms & 9 \\
        \addlinespace[0.5ex]
        Types of rooms & bathroom, bedroom 1, bedroom 2, dining room, dressroom, home gym, kitchen, living room, study \\
        \midrule
        \multirow{9}{*}{Objects} 
        & \textbf{bathroom}: toilet \\ \addlinespace[0.5ex]
        & \textbf{bedroom 1}: bed, pillows, plant, bulbs, floor lamp, window \\ \addlinespace[0.5ex]
        & \textbf{bedroom 2}: bed, pillow, floor lamp, window \\ \addlinespace[0.5ex]
        & \textbf{dining room}: table, chairs, plates, bowls, plant \\ \addlinespace[0.5ex]
        & \textbf{dressroom}: wardrobes, window, clothes \\ \addlinespace[0.5ex]
        & \textbf{home gym}: exercise machine, cycle \\ \addlinespace[0.5ex]
        & \textbf{kitchen}: wines, fridge, pot, microwave, sink, cupboard, dish wahser \\ \addlinespace[0.5ex]
        & \textbf{living room}: sofa, stand, vase, table, TV, window \\ \addlinespace[0.5ex]
        & \textbf{study}: desk, books, chair, lamp, computer, bookshelf, window \\
        \midrule
        Trajectory & living room $\rightarrow$ bathroom $\rightarrow$ bedroom 1 $\rightarrow$ kitchen $\rightarrow$ dining room $\rightarrow$ bedroom 2 $\rightarrow$ dressroom $\rightarrow$ study $\rightarrow$ home gym \\
        \midrule
        \multirow{3}{*}{Room Relations}
        & \textbf{Adjacency}: bedroom 1--bathroom--living room, dining room--kitchen \\ \addlinespace[0.5ex]
        & \textbf{Opposite}: study--dressroom, dressroom--bedroom 2, bedroom 1--dining room, living room--kitchen  \\ \addlinespace[0.5ex]
        & \textbf{Vertical}: dressroom--bedroom 1, living room--bedroom 2, dining room--study \\
        \bottomrule
    \end{tabularx}
    \caption{Scenario 10}
    \label{tab:scenario10}
\end{table}
\clearpage
\clearpage

\end{document}